\documentclass[11pt]{article}
\usepackage[utf8]{inputenc}
\usepackage{algorithm,algorithmic}
\usepackage{amsfonts}
\usepackage{amsmath}
\usepackage{amsthm}
\usepackage{color}
\usepackage{newtxtext}
\usepackage{graphicx}
\usepackage[margin=1in]{geometry}
\usepackage{mathtools}
\usepackage{multirow} 
\usepackage{booktabs}
\usepackage{bm}
\usepackage{authblk}
\usepackage{url}
\usepackage{calc}
\usepackage{enumitem}
\graphicspath{{figs/}}
\usepackage{caption}
\usepackage{subfigure}
\usepackage{hyperref}
\usepackage{rotating} 
\usepackage{pifont}
\usepackage{amssymb}
\usepackage{xltabular}
\title{Variational operator learning: A unified paradigm marrying training neural operators and solving partial differential equations}

\author[1]{Tengfei Xu}
\author[1]{Dachuan Liu}
\author[1]{Peng Hao \thanks{Corresponding author: haopeng@dlut.edu.cn}}
\author[1]{Bo Wang}
\affil[1]{State Key Laboratory of Structural Analysis, Optimization and CAE Software for Industrial Equipment, Department of Engineering Mechanics, Dalian University of Technology, Dalian 116024, China}
\date{}

\begin{document}
	\maketitle
	
	\begin{abstract}

		Neural operators as novel neural architectures for fast approximating solution operators of partial differential equations (PDEs), have shown considerable promise for future scientific computing. However, the mainstream of training neural operators is still data-driven, which needs an expensive ground-truth dataset from various sources (e.g., solving PDEs' samples with the conventional solvers, real-world experiments) in addition to training stage costs. From a computational perspective, marrying operator learning and specific domain knowledge to solve PDEs is an essential step for data-efficient and low-carbon learning. We propose a novel data-efficient paradigm that provides a unified framework of training neural operators and solving PDEs with the domain knowledge related to the variational form, which we refer to as the variational operator learning (VOL). We develop Ritz and Galerkin approach with finite element discretization for VOL to achieve matrix-free approximation of system functional and residual. We then propose direct minimization and iterative update as two possible optimization strategies. Various types of experiments based on reasonable benchmarks about variable heat source, Darcy flow, and variable stiffness elasticity are conducted to demonstrate the effectiveness of VOL. With a label-free training set and a 5-label-only shift set, VOL learns solution operators with its test errors decreasing in a power law with respect to the amount of unlabeled data. To the best of the authors' knowledge, this is the first study that integrates the perspectives of the weak form and efficient iterative methods for solving sparse linear systems into the end-to-end operator learning task.

	\end{abstract}
	
	\section{Introduction}
	\label{sec:introduction}
	
	Solving partial differential equations (PDEs) has widespread applications in science and engineering. Conventional solvers based on a variety of classical numerical approaches (e.g., finite element methods (FEMs), finite volume methods (FVMs), finite difference methods (FDMs), meshfree methods) have been developed and achieve great achievements in the last decades, and they can give solutions that meet engineering accuracy requirements under appropriate settings. Commercial software developed on the basis of these conventional solvers has been used extensively in engineering. However, the conventional solvers cannot learn knowledge or experience from their history of solving previous problems, which means they have to solve a new problem from scratch each time, even if it is highly similar to the ones they have solved before. The limitation leads to heavy computational costs in scenarios that require multiple simulations of different parameters, such as inverse problems, optimum design and uncertainty quantification. With the current rapid development of deep learning algorithms, software and hardware, training neural networks to learn the solution \emph{operators} \cite{Lu2021,li_2020_neural,li2020fourier,huang_2021_metaautodecoder,NOMAD,GOSWAMI2022114587,goswami_2022_deep,jin2022mionet,hao2023gnot,tran2023factorized} becomes a promising way to break through the above limitation of the conventional solver. But the training process of neural networks requires often a number of labels from simulation or experiments as ground truth data, which can be scarce and expensive to obtain. Many attempts \cite{Weinan2018,Sirignano2018,raissi_2019_physicsinformed,pmlr-v120-khodayi-mehr20a,samaniego_2020_an,Gao2021,Ranade2021,wang_2021_learning,li_2021_physicsinformed,GOSWAMI2022114587,rao_2023_encoding} (see Table \ref{tab:introductionsummary}) to embed domain knowledge like physical laws and principles into neural networks have been made to make the training process label-efficient and even label-free. Like the classical numerical approaches that solve PDEs with different forms, these techniques also deal with different forms of PDEs: the differential form (the strong form), the equivalent integral form, the variational form (the weak form) and the minimization form (see Table \ref{tab:introductionsummary}). The weak form transforms differential equations into integral equations, reducing the difficulty of solving by lowering the order of PDEs. Compared to the strong form, the weak form reduces the continuity requirements of the solution, which is less strict and more robust, making it appropriate for describing many of the real-world phenomena with non-smooth and even discontinuous physics. Besides, many powerful classical numerical approaches, especially FEMs and FVMs, are based on the weak form, which have developed mature mechanisms for handling boundary conditions and determining convergence of solutions. Embedding the weak form with these well-developed mechanisms in deep neural networks can help us overcome some of the difficulties like boundary condition management \cite{SUKUMAR2022114333} encountered in the strong form embedding. Moreover, the classical Ritz method and Galerkin method based on the weak form transform the original PDEs problem into solving a sparse linear system, which a lot of iterative techniques \cite{ysaad_2003_iterative} can be applied to solve with low costs. We provide Table \ref{tab:introductionsummary}, where we list of some of the representative existing literature of solving and learning PDEs, from classical approaches to some recent work combining machine learning. 
	From Table \ref{tab:introductionsummary}, we observe three such combinations, which are physics-informed DeepONet (PI-DeepONet) \cite{wang_2021_learning}, physics-informed neural operator (PINO) \cite{li_2021_physicsinformed} and physics-informed variational DeepONet (V-DeepOnet) \cite{GOSWAMI2022114587}. However, both PI-DeepONet and PINO are mostly based on the strong form. V-DeepONet, like the deep Ritz method \cite{Weinan2018} and the literature \cite{samaniego_2020_an}, chooses to set the system functional as the loss function, which is based on the minimization form. Combining operator learning model and the specific domain knowledge, especially the weak form, is worth exploring and will lead us to new numerical approaches suitable for solving parameterized PDEs. 
	
	% Motivated by the importance of the weak form and FEMs developed based on it, we marry the idea of operator learning and the weak form of PDEs with state-of-the-art neural operators and the variational method in this work.flexible
	% Some methods focus on strong form embedding to learn general functions \cite{raissi_2019_physicsinformed,rao_2023_encoding} and operators \cite{wang_2021_learning}.

	\paragraph{Our contributions.}

	In this Article, we propose the variational operator learning (VOL), a novel paradigm that combines training neural operators and solving partial differential equations with the variational form. To the best of the authors' knowledge, this is the first study that integrates the perspectives of the weak form and efficient iterative methods for solving sparse linear systems into the end-to-end operator learning task. The proposed VOL trains neural operators with a label-free training set. The distribution-shift operation with a 5-label-only shift set (see section \ref{sec:VOL}), which also exists in the conventional data-driven paradigm in the previous work \cite{li2020fourier}, is the only part requiring labels in the VOL algorithm. The main contributions of this work are summarized as follows:

	\begin{itemize}
 
		% A functional approximation with the node solution and the parameters of PDEs is derived and implemented with the tensor engine of deep learning. The variational operation is implemented with automatic differentiation. Thus a forward-backward propagation loop between the node solution and the functional approximation is developed, allowing calculating the residual of the system (i.e., calculating Eq. \ref{eq:residualerror}.) without acquiring the element stiffness and assembling the global stiffness matrix.

		\item Based on the idea of Ritz method and Galerkin method, we propose Ritz approach and Galerkin approach respectively in the framework of VOL. These two approaches can approximate system functional and system residual with FEM discretization in a matrix-free manner. 
		\item We introduce direct minimization and iterative update as two optimization strategies into the framework of VOL to minimize the residual norm. Specifically, for iterative update strategy, we integrate steepest decent method (SD) and conjugate gradient method (CG) into VOL with an efficient restart-update manner.
		\item   
		We investigate VOL with various experimental results. Our scaling experiments show the proposed VOL can learn operators and solve the PDEs effectively across different benchmarks given enough label-free data. We also conduct resolution experiments, comparative experiments verifying generalization benefits of VOL, and comparative experiments for different optimization strategies.

		% Specifically, in this Article, conjugate gradient (CG) is mainly used to provide efficient update steps for the VOL.
		% \item 
		% We propose that \emph{generalization} of neural operators can improve the performance of the iteration methods by providing a progressively better initial guess of solution for iteration methods in the framework of the proposed VOL. Experimental results show the generalization mechanism of neural operators in VOL can reduce the mean relative $L_2$ error by an order of magnitude.
		% \item
		% Cheaper than automatic differentiation: we derive derivatives with priori knowledge of discretization scheme.
		% \item
		% Our work connects neural network tensor operation with product of sparse matrix and vector, allowing Gauss quadrature, with Ritz and Galerkin approaches implemented, so that more complex numerical methods can be applied, such as Lanczos, CG and GMRES. Besides, we can design high-performance preconditioner, with approximate inverse. Feedback from iterative solver. Due to Ax is very economical, we can collect them inside the loop of the iterative solvers.
		
		% show that VOL can train neural operators on steady heat transfer and variable stiffness elasticity problems effectively with almost no labels. Five to ten labeled data are used in the distribution-shift session of the output. No data-driven loss term that requires external labels is used in all experiments in this Article.
		
	\end{itemize}
	
	\section{Related work}
	\label{sec:relatedwork}
	
	\paragraph{Surrogate modeling.}
	Surrogate modeling has been proposed to alleviate the computational burden of conventional solvers. Two main strategies, reduced order modeling methods (ROMs) and data-fit modeling methods (DFMs), have been adopted in the surrogate modeling. The main idea of these strategies is to seek a surrogate model of the original complex model so as to reduce or avoid the use of the conventional solver at the evaluation stage. ROMs \cite{hesthaven2016certified} use limited available snapshots from the original high-fidelity computational models to build simplified ones in the space of reduced basis. The constructed reduced-order models are computationally efficient compared with original full-order models. On the other hand, DFMs \cite{eldred2004second} including response surface methods, Kriging methods and neural networks, perform interpolation or regression of parameter-response pairs sampled from the high-fidelity dataset. The popular deep learning approach can also be considered as a kind of DFMs. An enormous amount of work of deep neural networks, including the neural operators \cite{Nie2020,Jiang2021,Gao2021,li_2020_neural,li2020fourier,Lu2021}, has focused on utilizing deep learning techniques to design excellent neural surrogate models.
	
	Two stages are usually required for both ROMs and DFMs to get a surrogate model in a purely data-driven manner: 

	(1) \emph{Data preparation}. At this stage, high-fidelity labels from the simulation of the conventional solvers or real-world experiments are produced and collected. Both ROMs and DFMs require often a fair number of such expensive labels.
	
	(2) \emph{Model construction (model training)}. Available labeled data from stage (1) are utilized to construct the surrogate model, which brings another computational cost. 
	
	It is worth noting that, despite the existence of techniques such as active learning \cite{settles.tr09} that tightly couple these two stages, they are in reality isolated from each other. From a model perspective, the data preparation stage just provides limited labels as examples, and keeps itself a black box to the surrogate models to be trained in the whole process of simulating or experimenting. Besides, from a solver perspective, the model training stage has no influence on the solving process in the data preparation stage, for example, it cannot accelerate the convergence of solving. Models trained in such a purely data-driven fashion can only acquire knowledge indirectly through the labels. If the model can learn directly from the domain knowledge, it is possible to skip the data preparation stage, and the model training process is to a certain extent equivalent to solve the original PDEs. Two ways to help coupling the model with domain knowledge, i.e., domain knowledge embedding and deep model embedding, are discussed in the next paragraph. 
	
	\paragraph{Domain knowledge embedding and deep model embedding.}

	Domain knowledge embedding is to embed domain knowledge (e.g., governing equations, discretization schemes, symmetries, variational principles) into the architecture of the (deep learning) model and the training process. We have listed some of the representative domain knowledge embedding methods in Table \ref{tab:introductionsummary}. Broadly speaking, domain knowledge embedding methods can either be mesh-free \cite{Weinan2018,Sirignano2018,raissi_2019_physicsinformed,pmlr-v120-khodayi-mehr20a,samaniego_2020_an,wang_2021_learning,GOSWAMI2022114587} or mesh-based \cite{Gao2021,Ranade2021,li_2021_physicsinformed}. In the mesh-free framework, residual of certain forms of PDEs is computed at some sampled positions in the solution domain, where the residual computation is heavily relied on automatic differentiation with the neural ansatz. On the other hand, in the mesh-based framework, the residual is computed with a certain discretization scheme, e.g., FDMs \cite{Gao2021}, FVMs \cite{Ranade2021}. We introduce them with some representative work:

	(1) \emph{Mesh-free framework with automatic differentiation.} 
	Automatic differentiation is a powerful tool to calculate derivatives in deep learning. The derivatives of the output with respect to input parameters such as spatial coordinates and time, the Jacobian matrix, and even the Hessian matrix can be easily obtained with automatic differentiation. Thus, the governing equations also their weak forms with derivative terms can be implemented elegantly in deep learning. Here we list some representative research that uses automatic differentiation to implement domain knowledge embedding. Deep Ritz method \cite{Weinan2018} constructs a functional with automatic differentiation approximating derivatives of the trial function, and numerical integration, which sets the minimization objective as the functional. Deep Galerkin method (DGM) \cite{Sirignano2018} and physics-informed neural network (PINN) \cite{raissi_2019_physicsinformed} are very similar, both of which use the residual error of the strong form as part of the loss function, deriving the residual term with automatic differentiation. Besides, both of them also treat the boundary conditions as penalty terms. PI-DeepONet \cite{wang_2021_learning} also calculates derivatives in loss function with automatic differentiation. Note that treating the strong form residual as a loss function can be viewed as embedding an equivalent integral form obtained through the collocation method with the delta function as the test function in the loss function. However, we have not classified it as an equivalent integral form in Table \ref{tab:introductionsummary} for the sake of clarity.
	
	% It is noted that treating strong form residual as loss function can also be considered as embedding an equivalent integral form derived by the collocation method with the delta function as the test function in the loss function, but we have not categorized it as equivalent integral form in Table \ref{tab:introductionsummary} for the sake of clarity. 
	
	(2) \emph{Mesh-based framework with a certain discretization scheme.}
	Discretization technique, on the other hand, computes the derivatives and the residual of PDEs with various discretization schemes. Tricks about convolution filters have been adopted to represent various discretization schemes. Motivated by the domain knowledge of the multigrid algorithm, a deep neural architecture called MgNet \cite{he_2019_mgnet} is designed and successfully used in dataset CIFAR-10 and CIFAR-100 \cite{Krizhevsky2009LearningML} for image classification. The following Meta-MgNet \cite{CHEN2022110996} use convolutions to represent discrete forms of differential operators, providing improved smoothers for the multi-grid algorithm. FEA-Net \cite{Yao2020} uses convolution kernel to express stiffness feature of the structure discretized by finite elements. PhyGeoNet \cite{Gao2021} and DiscretizationNet \cite{Ranade2021} discretize the output of the deep learning model with the finite difference discretization scheme and finite volume discretization scheme respectively to approximate the derivative terms and the residual in the governing equations. The proposed VOL in this Article chooses FEMs as the discretization scheme. For VOL, we design Ritz approach and Galerkin approach (see section \ref{sec:matrixfree})to form the global algebraic equation system in FEMs and calculate the residual in a matrix-free manner. Compared the mesh-free framework, Galerkin approach in VOL derives residual with no automatic differentiation, which is more resource-efficient in terms of computation and video memory. Compared with existing domain knowledge embedding work that also utilizes FEMs as the discretization scheme, for example, Meta-MgNet \cite{CHEN2022110996}, which reduces its Q1-element implementation to FDM, FEA-Net \cite{Yao2020}, which derives its convolution kernel analytically, VOL considers standard isoparametric elements. Moreover, VOL is allowed to perform Gauss quadrature of arbitrary order and shape function of arbitrary order, which is more practical and closer to actual engineering. 
	% Compared with existing domain knowledge embedding work that also utilizes FEMs as the discretization scheme, like Meta-MgNet \cite{CHEN2022110996} and FEA-Net \cite{Yao2020}, our approach considers isoparametric elements and Gauss quadrature, which is closer to actual engineering. 

	In contrast to domain knowledge embedding, deep model embedding is to embed the deep learning model in the classical numerical approach to enhance the capabilities of the classical numerical approaches. Deep potential \cite{han_2018_deep} leverages deep neural network representation of the potential energy surface for atoms and molecules system, which is a promising alternative to the classical potential representation in molecular dynamics and Monte Carlo simulations. Problem-independent machine learning \cite{Huang2022} embeds a simple feedforward neural network in the framework of extended multi-scale finite element method \cite{Zhang2010} to learn the mapping between discretized material density field of the coarse element and the multi-scale numerical shape functions of the element. Deep conjugate direction method \cite{Kaneda2022} uses the output of a deep convolutional neural network embedded in the algorithm to construct a good search direction which accelerates the convergence for solving large, sparse, symmetric, positive-definite linear systems. Fourier neural solver \cite{Cui2022} embeds a neural network in the stationary iterative method to help to eliminate error components in frequency space. 
	
	For both approaches, we just list some of the representative research in recent years. It is also noted that these two approaches are not mutually exclusive, instead, they are complementary to each other, which means a method can be both domain knowledge embedding and deep model embedding, and we can use either one or both of them to solve the problem.
	
	% However, by marrying the domain knowledge and the deep learning models, we  can be achieved than using deep learning model alone and using classical numerical approaches alone. Specifically, domain knowledge embedding can help reduce label use of deep learning models \cite{} or design new neural architectures \cite{} with embedding domain knowledge into the deep learning models, and deep model embedding enhances the capabilities of the the classical numerical approach by the various useful properties of the deep learning model. they can be complementary to each other
	% In summary, domain knowledge embedding and deep model embedding do not necessarily reduce or eliminate the usage of labels. 

	\paragraph{Operator learning and neural operators.}
	Operator learning is to let the model learn the operator between two function spaces. Operator learning models, which we say are \emph{operator\text{-}based}, can give prediction over a whole parameter set of parameterized PDEs. In Table \ref{tab:introductionsummary}, we classify some of the representative existing methods according to whether they are function-based or operator-based. Neural operators refer to specific neural network architectures designed for operator learning. According to the type of input and output of the neural networks, the existing architectures of neural operators can be divided into two categories: 
	
	(1) \emph{Point-wise}. Inspired by the universal approximation theorem for operators \cite{chen1995approximation,chen1995universal}, deep operator network (DeepONet) \cite{Lu2021}, as a representative architecture of this category, receives the parameter field and a query point, and then output solution of the query points in the computational domain, which has a branch net for encoding discrete function space and a trunk net for encoding the coordinate information of query points, and the output of which is the inner product of the output of two nets. Following work based on DeepONet includes learning multiple-input operators \cite{jin2022mionet}, combing DeepONet with physics-informed machine learning \cite{wang_2021_learning}, replacing the trunk net with basis functions precomputed by proper orthogonal decomposition \cite{lu_2022_a}. Recently, a general neural operator transformer (GNOT) \cite{hao2023gnot} is proposed, with a heterogeneous normalized attention layer design, and GNOT is also designed to handle multiple input functions and irregular meshes.
	
	(2) \emph{Field-wise}. These architectures \cite{li_2020_neural,li2020fourier,you_2022_learning,tran2023factorized} input discrete parameter fields and output discrete solution fields. Fourier neural operators (FNO) \cite{li2020fourier} parameterize the integral kernel in the Fourier space and utilize the idea of shortcut connection, producing a powerful Fourier layer. The implicit Fourier neural operator (IFNO) \cite{Gao2021} utilizes the Fourier layer in an implicit manner, which lets the data flow pass through the Fourier layer recurrently and has better training stability. Factorized Fourier neural operator (F-FNO) \cite{tran2023factorized} adopts Fourier factorization and a handful of other techniques about network design and training settings to enhance the model performance. In this Article, we focus on training the field-wise neural operators due to the natural similarity between field-wise output and mesh settings in FEMs. 
	
	It is worth noting that, in addition to these neural operators specifically designed for operator learning, the idea of operator learning has also been combined with various neural architectures and approaches, including convolutional neural networks \cite{Nie2020} and the generative neural networks \cite{Jiang2021}, principal component analysis \cite{bhattacharya_2021_model,lanthaler_2023_operator}, meta-learning \cite{huang_2021_metaautodecoder}, transfer learning \cite{goswami_2022_deep}, attention mechanism \cite{georgioskissas_2022_learning,hao2023gnot}, manifold learning \cite{NOMAD}. 

	% Meta learning MAD and NOMAD transformer and coupled attention
	% Compared to \emph{function\text{-}based} models that can only give prediction of a single parameter, operator learning models have enormous potential in applications like inverse problems, optimum design and uncertainty quantification.transfer learning (deep transfer operator learning)
	
	% \paragraph{Neural accelerated iterative solvers.}

	% Or named as neural solvers.
	% papers: 
	% Meta-MgNet 
	% Preconditioners:
	% Multigrid-aug preconditioner: matrix-free finite difference stencil 
	% Deep Learning of Preconditioners for Conjugate Gradient Solvers in Urban Water Related Problems learn matrix, not matrix-free
	% HINTS: learn the update
	% We learn the inverse of a family of coefficient matrices
	
	\begin{figure}[htbp]
		\centering
		\includegraphics[width=\textwidth]{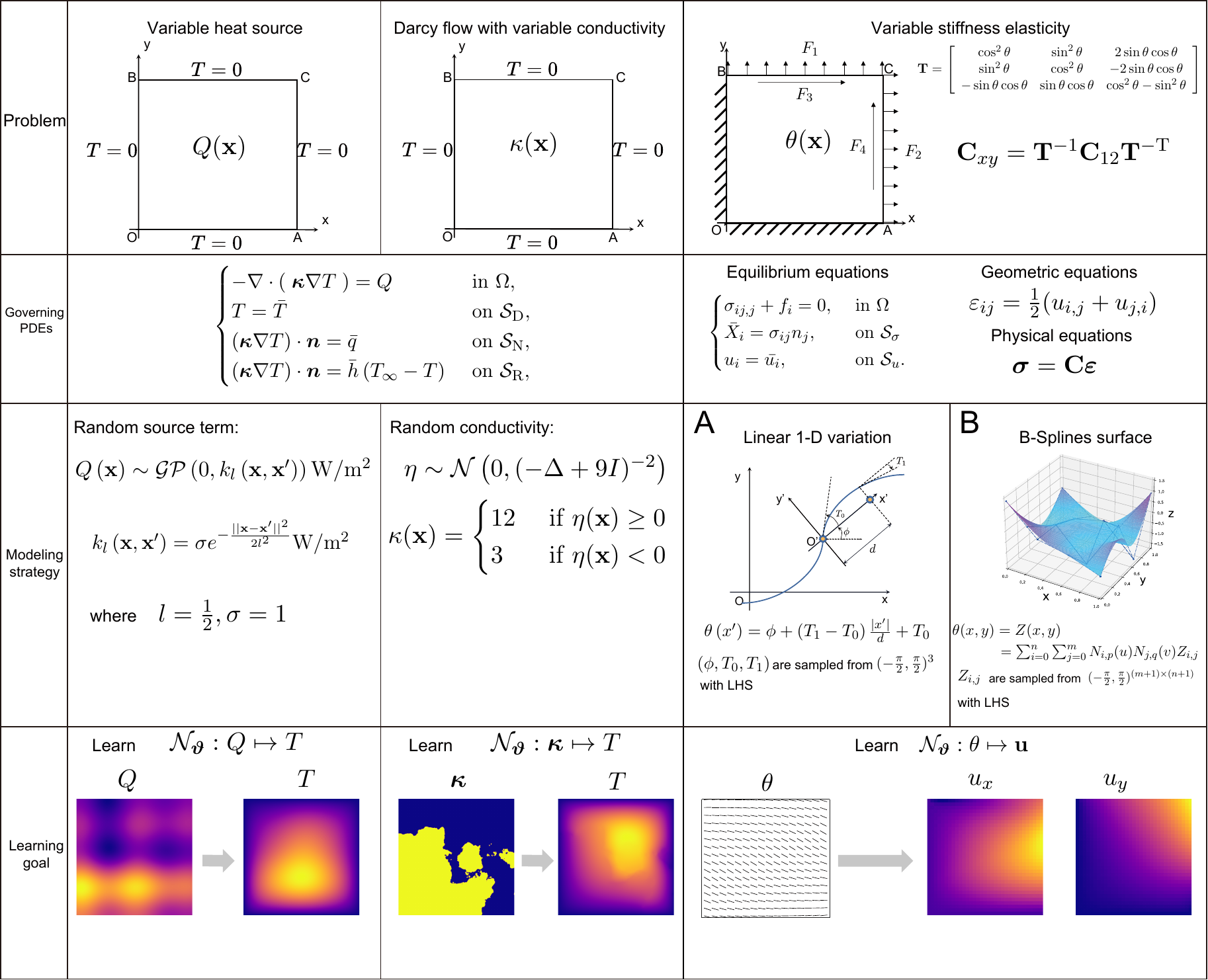}
		\caption{A schematic representation of the VOL benchmarks in this work.}
		\label{fig:cases}
	\end{figure}

	% \begin{figure}[htbp]
	% 	\centering
	% 	\includegraphics[width=\textwidth]{formofPDEs.pdf}
	% 	\caption{Different forms of PDEs.}
	% 	\label{fig:formsofpdes}
	% \end{figure}
	
	\begin{sidewaystable}[htbp]
		\centering
		\resizebox{\linewidth}{!}{\begin{tabular}{cccccccccc}
				\hline
				\multirow{2}{*}{}          & \multirow{2}{*}{Function-based} & \multirow{2}{*}{Operator-based} & \multicolumn{3}{c}{Type of methods}                                                  & \multicolumn{4}{c}{Form of PDEs}                                                                              \\ \cmidrule(r){4-6} \cmidrule(l){7-10}
				&                                &                                & Classical numerical approach & Data-driven modeling & Domain knowledge embedding & Differential form               & Equivalent integral form  & Minimization form           & Variational form                 \\ \hline
				Analytical methods         & \checkmark      &      \checkmark                          &     &                           &                           & \checkmark &                           &                           &                           \\
				FDMs                       & \checkmark      &                                & \checkmark    &                           &                           & \checkmark &                           &                           &                           \\
				Weighted residual methods  & \checkmark      &                                & \checkmark    &                           &                           &                           & \checkmark &                           &                           \\
				Ritz method                & \checkmark      &                                & \checkmark    &                           &                           &                           &                           &  &  \checkmark                         \\
				Galerkin method                & \checkmark      &                                & \checkmark    &                           &                           &                           &                           &  &  \checkmark                         \\
				FEMs                       & \checkmark      &                                & \checkmark    &                           &                           &                           &                           &                           & \checkmark \\
				FVMs                       & \checkmark      &                                & \checkmark    &                           &                           &                           &                           &                           & \checkmark \\
				ROMs                       &     &     \checkmark                             &     &         \checkmark                  &                           &                           &                           &                           &  \\
				Deep Ritz method \cite{Weinan2018}      & \checkmark      &                                &                              &                           & \checkmark &                           &                           & \checkmark &                           \\
				DGM \cite{Sirignano2018} & \checkmark      &                                &                              &                           & \checkmark & \checkmark &                           &                           &                           \\
				PINN \cite{raissi_2019_physicsinformed}                  & \checkmark      &                                &                              &                           & \checkmark & \checkmark &                           &                           &                           \\
				VarNet \cite{pmlr-v120-khodayi-mehr20a}           & \checkmark      &                                &                              &                           & \checkmark &                           &                           &  &    \checkmark                       \\				
				Literature \cite{samaniego_2020_an}           & \checkmark      &                                &                              &                           & \checkmark &                           &                           & \checkmark &                           \\
				PhyGeoNet \cite{Gao2021}             &                                & \checkmark      &                              &                           & \checkmark & \checkmark &                           &                           &                           \\
				DiscretizationNet \cite{Ranade2021}     & \checkmark      &                                &                              &                           & \checkmark &                           &                           &                           & \checkmark \\
				DeepONet \cite{Lu2021}              &                                & \checkmark      &                              & \checkmark &                           &                           &                           &                           &                           \\
				FNO \cite{li2020fourier}                   &                                & \checkmark      &                              & \checkmark &                           &                           &                           &                           &                           \\
				PI-DeepONet \cite{wang_2021_learning}           &                                & \checkmark      &                              &                           & \checkmark & \checkmark &                           &                           &                           \\
				PINO \cite{li_2021_physicsinformed}                  &                                & \checkmark      &                              &                           & \checkmark & \checkmark &                           &       \checkmark                     &                           \\ 
				V-DeepONet \cite{GOSWAMI2022114587}                  &                                & \checkmark      &                              &                           & \checkmark & &                           &              \checkmark              &                           \\				
				\textbf{VOL (ours)}                 &                                & \checkmark      &                              &                           & \checkmark &                           &                           &                           & \checkmark \\ \hline
				
		\end{tabular}}
		
		\caption{{Summary of some of the representative existing methods of solving and learning PDEs.}}
		\label{tab:introductionsummary}
	\end{sidewaystable}

	\section{Preliminaries}
	\label{sec:Preliminaries}
	
%In this section, we introduce four forms of PDEs. We treat collocation method as strong form. Ritz and Galerkin method and our corresponding implementation are introduced respectively. 

%Let V be a Hilbert space with the inner product $a\left(\cdot,\cdot \right):V \times V \to \mathbb{R}$ and norm $\|\mathbf{v}\|_V=a(\mathbf{v}, \mathbf{v})^{1 / 2}$.
	\subsection{Forms of partial differential equations}
\label{sec:formspdes}
Consider $\Omega \subseteq \mathbb{R}^d$, where $d$ is a fixed number in $\mathbb{N}^+$. $\mathcal{S}$ denotes the boundary of $\Omega$. Without loss of generality, we start with the strong form of the stationary PDEs:

% BVPs

\begin{equation}\label{eq:strongform}
	\begin{gathered}
		\mathbf{A}\left(\mathbf{u}\left(\mathbf{x}\right)\right)=0 , \mathbf{x} \in \Omega\\
		\mathbf{BC}\left(\mathbf{u}\left(\mathbf{x}\right)\right)=0, \mathbf{x} \in\mathcal{S}
	\end{gathered}
\end{equation}

Multiply Eq. \ref{eq:strongform} by an arbitrary test function $\mathbf{v} \in \mathcal{C}_{\mathrm{c}}^{\infty}(\Omega)$, we get the weighted integral form or equivalent integral form of PDEs

\begin{equation}
	\int_{\Omega} \mathbf{v}^T \mathbf{A}(\mathbf{u}) \mathrm{d} \Omega+\int_{\mathcal{S}} \mathbf{v}^T \mathbf{BC}(\mathbf{u}) \mathrm{d} \mathcal{S} \equiv 0.
\end{equation}

Then, with integration by parts lowering the order of variables, we can get the weak form of PDEs

\begin{equation}\label{eq:weakform}
\int_{\Omega} \mathbf{C}(\mathbf{v})^\mathrm{T} \mathbf{D}(\mathbf{u}) \mathrm{d} \Omega+\int_{\mathcal{S}} \mathbf{E}(\mathbf{v})^\mathrm{T} \mathbf{F}(\mathbf{u}) \mathrm{d} \mathcal{S} \equiv 0. 
\end{equation}

And note the weak form can be expressed in such an abstract form

\begin{equation}\label{eq:weakabstractform}
	\begin{gathered}
\text{Find } \mathbf{u} \in V \text{, such that } a\left(\mathbf{u}, \mathbf{v}\right) = L\left( \mathbf{v}\right), 
	\end{gathered}
\end{equation}
When $a\left(\mathbf{u}, \mathbf{v}\right)$ is symmetric, we can also write the minimization form of the PDEs. By treating the test function $\mathbf{v}$ as the variation of $\mathbf{u}$, we can derive the functional $\Pi$ of the system from the weak form. In this case, the original problem (solving PDEs) has been transformed into a functional minimization problem

\begin{equation}\label{eq:energyform}
	\begin{gathered}
	\Pi=\Pi\left(\mathbf{u} \right) \to\min_{\mathbf{u}} \Pi (\mathbf{u}).
	\end{gathered}
\end{equation}

Note $\Pi$ can also be written in such an abstract form

\begin{equation}\label{eq:energyabstractform}
	\begin{gathered}
		\Pi= \frac{1}{2}a(\mathbf{u}, \mathbf{u})-L\left( \mathbf{u}\right). 
	\end{gathered}
\end{equation}

The solutions to Eq. \ref{eq:weakform}, Eq. \ref{eq:weakabstractform} and Eq. \ref{eq:energyform}, Eq. \ref{eq:energyabstractform} are called \emph{weak solutions}. Eq. \ref{eq:energyform} and Eq. \ref{eq:energyabstractform} are also called "weak form" sometimes, but to emphasize the Eq. \ref{eq:energyform} and Eq. \ref{eq:energyabstractform} minimize the system functional, while the weak form provides a more general approach, we call them minimization form in this Article.

% in this Article, what we want to emphasize is, the Eq. \ref{eq:energyform} Eq. \ref{eq:energyabstractform} don't exist anymore when $a\left(\mathbf{u}, \mathbf{v}\right)$ is symmetric, while the weak form still holds .

%Here we introduce two key methods closely related to our work: Ritz method and Galerkin method. Ritz method 

	\subsection{Ritz method and Galerkin method}
\label{sec:ritzclassic}

Ritz method and Galerkin method are two representative ways of utilizing the weak form and minimization form to solve PDEs. They both approximate weak solutions in finite-dimensional spaces. We introduce these two methods with stationary problems of PDEs in Hilbert spaces.

\begin{itemize}
	
	\item\emph{Ritz method}
	
%	Ritz method focuses on minimizing functional in Eq. \ref{eq:energyform} and Eq. \ref{eq:energyabstractform} of the system. Choose $V_n$ a finite dimensional subspace of $V$, $dim(V_n)=n$, and then construct a basis $\left(\ \bm{\phi}_1,\cdots,\bm{\phi}_n\right) $ of $V_n$. The basis should be carefully chosen to satisfy the essential boundary condition, otherwise the , i.e., the stiffness matrix. Then, $\mathbf{u}$ can be approximated with the decomposition $\mathbf{u}^n$
%	
%	\begin{equation}\label{eq:ritzdecomposition}
%		\begin{gathered}
%		\mathbf{u}^n= \sum_{j=1}^{n}\mathbf{u}_j\bm{\phi}_j
%		\end{gathered}
%	\end{equation}
%	
%	Then, $\Pi$ can be written as
%	
%	\begin{equation}\label{eq:ritzPi}
%\begin{aligned}
%	\Pi\left(\mathbf{u}^n\right) & =\frac{1}{2} a\left(\mathbf{u}^n, \mathbf{u}^n\right)-L\left(\mathbf{u}^n\right) \\
%	& =\frac{1}{2} a\left(\sum_{j=1}^n \mathbf{u}_j \bm{\phi}_j, \sum_{i=1}^n \mathbf{u}_i \bm{\phi}_i\right)-L\left(\sum_{i=1}^n \mathbf{u}_i \bm{\phi}_i\right) \\
%	& =\frac{1}{2} \sum_{i=1}^n \sum_{j=1}^n a\left(\mathbf{u}_j \bm{\phi}_j, \mathbf{u}_i \bm{\phi}_i\right)-\sum_{i=1}^n L\left(\mathbf{u}_i \bm{\phi}_i\right) \\
%	& =\frac{1}{2} \sum_{i=1}^n \sum_{j=1}^n \mathbf{u}_j \mathbf{u}_i a\left(\bm{\phi}_j, \bm{\phi}_i\right)-\sum_{i=1}^n \mathbf{u}_i L\left(\bm{\phi}_i\right)
%\end{aligned}
%	\end{equation}
%The basis should be carefully chosen to satisfy the essential boundary condition, otherwise the coefficient matrix, i.e., the global stiffness matrix, will end up being a singular one

	Ritz method focuses on minimizing functional in Eq. \ref{eq:energyform} and Eq. \ref{eq:energyabstractform} of the system. Choose $V_n$ a finite dimensional subspace of $V$, $dim(V_n)=n$, and then construct a basis $\left(\ \bm{\phi}_1,\cdots,\bm{\phi}_n\right) $ of $V_n$. Then, $\mathbf{u}$ can be approximated with the $\mathbf{u}^n$ uniquely decomposed on the basis

\begin{equation}\label{eq:ritzdecomposition}
	\begin{gathered}
		\mathbf{u}^n = \sum_{j=1}^{n}u_j\bm{\phi}_j.
	\end{gathered}
\end{equation}

Then, $\Pi$ can be written as

\begin{equation}\label{eq:ritzPi}
	\begin{aligned}
		\Pi\left(\mathbf{u}^n\right) & =\frac{1}{2} a\left(\mathbf{u}^n, \mathbf{u}^n\right)-L\left(\mathbf{u}^n\right) \\
		& =\frac{1}{2} a\left(\sum_{j=1}^n u_j \bm{\phi}_j, \sum_{i=1}^n u_i \bm{\phi}_i\right)-L\left(\sum_{i=1}^n u_i \bm{\phi}_i\right) \\
		& =\frac{1}{2} \sum_{i=1}^n \sum_{j=1}^n a\left(u_j \bm{\phi}_j, u_i \bm{\phi}_i\right)-\sum_{i=1}^n L\left(u_i \bm{\phi}_i\right) \\
		& =\frac{1}{2} \sum_{i=1}^n \sum_{j=1}^n u_j u_i a\left(\bm{\phi}_j, \bm{\phi}_i\right)-\sum_{i=1}^n u_i L\left(\bm{\phi}_i\right).
	\end{aligned}
\end{equation}	
	
Thus, $\Pi$ can be rewritten under algebraic form

	\begin{equation}\label{eq:ritzPialgebraicform}
	\begin{aligned}
		\Pi = \frac{1}{2}\mathbf{u}^{\mathrm{T}} \mathbf{A u}-\mathbf{u}^{\mathrm{T}} \mathbf{b},
	\end{aligned}
\end{equation}

where ${A}_{ij}=a\left(\bm{\phi}_j,  \bm{\phi}_i\right)$, ${b}_i=L\left(\bm{\phi}_i\right)$. To minimize $\Pi$, Ritz method lets the gradient of the function $\Pi\left(u_1,\cdots,u_n \right) $ be a zero vector
\begin{equation}
	\begin{gathered}
	\frac{\partial \Pi}{\partial \mathbf{u}}=\mathbf{A}\mathbf{u}-\mathbf{b}=\mathbf{0}.
	\end{gathered}
\end{equation}

%, which is equivalent to the first variation of the functional $\delta\Pi$ equals zero.

	\item\emph{Galerkin method}

Galerkin method is based on the weak form of PDEs. Like Ritz method, Galerkin method also uses the basis $\left(\ \bm{\phi}_1,\cdots,\bm{\phi}_n\right) $ of $V_n$ to construct approximation of $\mathbf{u}$. Based on Eq. \ref{eq:weakabstractform} and Eq. \ref{eq:ritzdecomposition}, we get the abstract form of Galerkin method

\begin{equation}\label{eq:galerkinabstract}
	\begin{gathered}
		\text{Find }\mathbf{u}^n \in V_n, \text{such that: }a\left(\mathbf{u}^n, \mathbf{v}_i\right)=L\left(\mathbf{v}_i\right). 
	\end{gathered}
\end{equation}

By assuming finite number of test functions $\mathbf{v}_1, \mathbf{v}_2,...\mathbf{v}_n$, Galerkin method turns the abstract form into a set of linear equations that can be solved numerically

\begin{equation}\label{eq:galerkinalgebraicform}
	\begin{aligned}
		\mathbf{A u} = \mathbf{b},
	\end{aligned}
\end{equation}

where ${A}_{ij}=a\left(\bm{\phi}_j, \mathbf{v}_i\right)$, ${b}_i=L\left(\mathbf{v}_i\right)$. The test functions can be either in $V_n$ or not, and the type of Galerkin method where $\mathbf{v}$ is not taken in $V_n$ is called Petrov-Galerkin method. Galerkin method is more general than Ritz method. Compared to Ritz method, Galerkin method does not require the symmetry of the bilinear form, allowing it to handle problems where the minimization form of PDEs does not exist.

%If the test function also in $V_n$, To transform Eq. \ref{eq:galerkinabstract} into algebraic form, every element of the basis of $V_n$ is chosen as $\mathbf{v}^n$, producing $n$ .
 	
\end{itemize}

Inspired by these two classical numerical methods, we develop Ritz approach and Galerkin approach of VOL respectively to approximate system functional and residual with deep learning toolkit (see section \ref{sec:matrixfree} and Fig. \ref{fig:ritzandgalerkin}).

\subsection{Finite element methods}
\label{sec:fem}

FEMs is a special case of Ritz method and Galerkin method, where finite subdomains (elements) are designed to be supports of trial functions and test functions. In this Article, VOL utilizes the discretization scheme of FEMs, and handles the global algebraic equation system of FEMs in a matrix-free and iterative manner (see section \ref{sec:matrixfree}).

%To handle such
%	introduce four forms. strong forms also can be seen as a weighted residual method.
% Note both ritz and galerkin do not use the minimization minimization form.	
% direct bminimization of energy will be bad in the discrete	system

\section{Results}
\label{sec:result and discussion}

Various experimental results are shown and discussed in this section to demonstrate the effectiveness the proposed VOL. Specifically, we investigate VOL on steady heat transfer problem with variable heat source, Darcy flow with variable conductivity, and variable stiffness elasticity (Fig. \ref{fig:cases}). We first introduce our problem settings briefly. We conduct scaling experiments to investigate the influence of the different data sizes on the performance of VOL. We then conduct the resolution experiments to verify the capability of VOL at different resolutions. We then design two comparative experiments between VOL and iterative methods to verify the generalization benefits of VOL. We also conduct experiments to compare the proposed optimization strategies in VOL with the data-driven strategy.

% In this section, we first investigate different optimization strategies for VOL, and choose the setting of the section \ref{sec:heatcase1}, \ref{sec:heatcase2}, \ref{sec:vspcase1} and \ref{sec:vspcase2} depending on the results of the investigation. Then, in section \ref{sec:heatcase1}, \ref{sec:heatcase2}, \ref{sec:vspcase1} and \ref{sec:vspcase2}, we demonstrate the capability of the proposed VOL for learning operators of the governing PDEs of steady heat transfer and variable stiffness elasticity. At last, we elaborate generalization benefits in VOL from the perspective of neural solver, and observe the generalization benefits with experimental results. 

% In VOL, we keep the same mesh settings with the corresponding commercial FEM solvers.
% on the darcy flow problem and variable elasticity problem 
% All experiments in this Article are implemented with Pytorch \cite{PyTorch2019}. Code and datasets for all experiments in this Article will be published at \href{https://github.com/BraveDrXuTF/VOL}{this repository}.

% Specifically, we learn Fourier neural operators on steady heat transfer problem with variable heat source Darcy flow, and variable elasticity 
\subsection{Problem settings}\label{sec:problemsetting}

\subsubsection{Variable stiffness elasticity}\label{sec:elasticproblemsetting}

% \begin{figure}[htbp]
% 	\centering
% 	%	\def\svgscale{0.75}
% 	\includegraphics[width=0.4\textwidth]{elasticbody.pdf}
% 	%	\input{elasticbody.pdf_tex}
% 	\caption{A linear elastic body with body forces and boundary conditions.}
% 	\label{fig:elasticbody}
% \end{figure}

\begin{itemize}\label{itm:elasticity}
	\item\emph{General elasticity}
	
	Consider a linear elastic body $\mathcal{B} \subset \mathbb{R}^3$ (Extended Data Fig. \ref{fig:elasticbody}a). Let the boundary of the body be $\mathcal{S}=\mathcal{S}_{\sigma}\cup \mathcal{S}_{u}$. The governing equations and boundary conditions of $\mathcal{B}$ are
	\begin{equation}\label{eq:elasticbody}
		\begin{cases}\sigma_{i j, j}+f_i=0, & \text { in } \Omega \\ \bar{X}_i=\sigma_{i j} n_j, & \text { on } \mathcal{S}_\sigma
			\\
			u_i=\bar{u_i}, & \text { on } \mathcal{S}_{u}.
		\end{cases}
	\end{equation}
	The weighted integral form of Eq. \ref{eq:elasticbody} is
	%where $\mathcal{\sigma}$ is the Cauchy stress tensor, $f=\{X, Y, Z\}^{\mathrm{T}}$ are the body forces, and $\bar{X}=\left\{\bar{X}, \bar{Y}, \bar{Z}\right\}^{\mathrm{T}}$ are the traction forces. The weighted integral form of Eq. \ref{eq:elasticbody} is
	\begin{equation}\label{eq:weightedint}
		\begin{gathered}
			\int_\Omega\left(\sigma_{i j, j}+f_i\right) \delta u_i\mathrm{d} \Omega+\int_{\mathcal{S}_\sigma}\left(\bar{X}_i-\sigma_{i j} n_j\right) \delta u_i\mathrm{d} \mathcal{S}=0, 
			\quad
			\forall \delta u_i \in\left\{v \in C[\Omega] ; v=0 \text { on } \mathcal{S}_u\right\}.
		\end{gathered}
	\end{equation}
	According to Green's formula,
	\begin{equation}\label{eq:fenbujifen}
		\begin{aligned}
			& \int_\Omega \sigma_{i j, j} \delta u_i\mathrm{d} \Omega-\int_{\mathcal{S}_\sigma} \sigma_{i j} n_j \delta u_i\mathrm{d} \mathcal{S}=\int_\Omega \sigma_{i j, j} \delta u_i\mathrm{d} \Omega-\int_\mathcal{S} \sigma_{i j} n_j \delta u_i\mathrm{d} \mathcal{S}=-\int_\Omega \sigma_{i j} \delta u_{i, j}\mathrm{d} \Omega \\
			& =-\int_\Omega \frac{1}{2} \sigma_{i j}\left(\delta u_{i, j}+\delta u_{j, i}\right)\mathrm{d} \Omega=-\int_\Omega \sigma_{i j} \delta \varepsilon_{i j}\mathrm{d} \Omega,
		\end{aligned}
	\end{equation}
Substituting Eq. \ref{eq:fenbujifen} into Eq. \ref{eq:weightedint}, we have
	\begin{equation}\label{eq:xugong}
		\begin{gathered}
			\int_\Omega f_i \delta u_i\mathrm{d} \Omega+\int_{\mathcal{S}_\sigma} \bar{X}_i \delta u_i\mathrm{d} \mathcal{S}=\int_\Omega \sigma_{i j} \delta \varepsilon_{i j}\mathrm{d} \Omega,
			\\
			\forall \delta u_i \in\left\{v \in C[\Omega] ; v=0, \text { on } \mathcal{S}_u\right\}.
		\end{gathered}
	\end{equation}
	Eq. \ref{eq:xugong} is also known as \emph{the principle of virtual work}, which is a variational form of Eq. \ref{eq:elasticbody}. The left term of Eq. \ref{eq:xugong} is the virtual work done by the external forces, and the right term is the strain energy of the elastic body $\mathcal{B}$ (also the opposite of the virtual work done by the internal forces). Thus, we have
	\begin{equation}
		\begin{gathered}
			\delta W_{\mathrm{int}}+\delta W_{\mathrm{ext}}=-\delta\left[U+U_{\mathrm{e}}\right]=0,
		\end{gathered}
	\end{equation}
	where $\delta W_{\mathrm{int}}$, $\delta W_{\mathrm{ext}}$ are the virtual work done by the internal forces and the virtual work done by the external forces respectively. $U$ is the strain energy of the elastic body $\mathcal{B}$, and $U_{\mathrm{e}}$ is the potential energy of the external forces. Let the system functional be $\Pi$, and we have
	\begin{equation}\label{eq:VP}
		\begin{gathered}
			\delta \Pi=0,
			\\
			\Pi=U+U_{\mathrm{e}},
			\\
			\forall \delta u_i \in\left\{v \in C[\Omega] ; v=0, \text { on } \mathcal{S}_u\right\}.
		\end{gathered}
	\end{equation}
	Note Eq. \ref{eq:VP} is equivalent to Eq. \ref{eq:elasticbody}, and here we only give the derivation of Eq. \ref{eq:VP} from Eq. \ref{eq:elasticbody}.
	
	For convenience, we rewrite the stress tensor and strain tensor in their vector form, i.e., $$\bm{\sigma}=\left[\sigma_{x}, \sigma_{y}, \sigma_{z}, \tau_{yz}, \sigma_{xz}, \sigma_{xy}\right]^{\mathrm{T}},  
	\bm{\varepsilon}=\left[\varepsilon_{x}, \varepsilon_{y}, \varepsilon_{z}, \tau_{yz}, \varepsilon_{xz}, \varepsilon_{xy}\right]^{\mathrm{T}}.$$
	Then the strain energy of the elastic body and the potential energy of the external forces is
	\begin{equation}\label{eq:energyexpression}
		\begin{aligned}
			& U=\int_\Omega \frac{1}{2} \bm{\varepsilon}^{\mathrm{T}} \mathbf{C} \bm{\varepsilon} \mathrm{d} \Omega ,\\
			& U_{\mathrm{e}}=-\int_{\Omega} \mathbf{f}^{\mathrm{T}}\mathbf{u}  \mathrm{d} \Omega -\int_{\mathcal{S}_\sigma} \mathbf{\bar{X}}^{\mathrm{T}} \mathbf{u}  \mathrm{d} \mathcal{S}.
		\end{aligned}
	\end{equation}
	Thus, functional $\Pi$ is:
	\begin{equation}\label{eq:continuumtotal}
		\Pi=U+U_{\mathrm{e}}=\int_\Omega\frac{1}{2} \bm{\varepsilon}^{\mathrm{T}} \mathbf{C} \bm{\varepsilon} \mathrm{d} \Omega -\int_{\Omega} \mathbf{f}^{\mathrm{T}}\mathbf{u}  \mathrm{d} \Omega -\int_{\mathcal{S}_\sigma} \mathbf{\bar{X}}^{\mathrm{T}} \mathbf{u}  \mathrm{d} \mathcal{S},
	\end{equation}
	where $\mathbf{C}$ is the matrix of material properties. Note $\Pi$ is the total potential energy of the system. 

	% has physical implication
	In context of operator learning, for example, $\mathbf{C}$ can be a function of coordinates $\mathbf{C}\left( x,y,z\right) $ in random function spaces, and our goal is to learn an operator learning model, which outputs the displacement field $\mathbf{u}$ that lets $\delta\Pi=0$ hold as close as possible for each input instance with $\mathbf{C}\left( x,y,z\right) $ in the test set. In our experiments, we consider a 2-D variable stiffness case, where $\mathbf{C}=\mathbf{C}(\theta(x,y))$.

	\item\emph{Variable stiffness elasticity} 

	An elastic variable stiffness square plate with in-plane deformation is considered in our experiments, and it is considered to be made of the fiber reinforced material, as shown in Fig. \ref{fig:cases}. The thickness of the plate is 0.125mm. Due to spatial variation of fiber orientation, the material property of the plate shows anisotropy. The material property matrix in x-y coordinate system $\mathbf{C}_{xy} $ can be written as:
\begin{equation}\label{eq:Cxy}
	\begin{gathered}
		\mathbf{C}_{xy}=\mathbf{T}^{-1}\mathbf{C}_{12}\mathbf{T}^{-\mathrm{T}}
	\end{gathered}
\end{equation}	
where
\begin{equation}\label{eq:Tmatrix}
	\begin{gathered}
		\mathbf{T}=\left[\begin{array}{ccc}
			\cos ^{2} \theta & \sin ^{2} \theta & 2 \sin \theta \cos \theta \\
			\sin ^{2} \theta & \cos ^{2} \theta & -2 \sin \theta \cos \theta \\
			-\sin \theta \cos \theta & \sin \theta \cos \theta & \cos ^{2} \theta-\sin ^{2} \theta
		\end{array}\right]
	\end{gathered}
\end{equation}	
And $\mathbf{C}_{12}$ is the material property matrix in the principle material coordinates, which is not effected by the fiber angle. Eq. \ref{eq:S12} gives the formulation of compliance matrix $\mathbf{S}_{12}$, i.e., the inverse of $\mathbf{C}_{12}$:
\begin{equation}\label{eq:S12}
	\begin{gathered}
		\mathbf{S}_{12}=\left[\begin{array}{lll}
			\frac{1}{E_1} & -\frac{\nu_{12}}{E_1} & 0 \\
			-\frac{\nu_{12}}{E_1} & \frac{1}{E_2} & 0 \\
			0 & 0 & \frac{1}{G_{12}}
		\end{array}\right]
	\end{gathered}
\end{equation}

The fiber angle field $\theta=\theta\left(x, y \right)$ of the plate is characterized in two ways as problem settings: (1) Elasticity A: Linear 1-D variation \cite{ZG1993,ZG2008}. (2) Elasticity B: B-splines surface \cite{Zhang2023}. The goal of VOL is to learn the mapping between the fiber angle field space and the vector space of displacement components $\left[u_1,u_2\right]^{\mathrm{T}}$. Specifically, we study a 100mm$\times$100mm fiber-reinforced panel. For details about problem settings of variable stiffness elasticity, including material properties, boundary conditions, elements and parameter range, see Fig \ref{fig:cases} and Supplementary material S2.2.

\end{itemize}

\subsubsection{Steady heat transfer with variable heat source and Darcy flow}\label{sec:heatproblemsetting}

Steady heat transfer with variable heat source in 1m$\times$1m$\times$1m cube and Darcy flow with variable conductivity in 1$\times$1 square are studied respectively (See Fig. \ref{fig:cases}). The governing equation of both steady heat transfer and Darcy flow is described as
\begin{equation} \label{eq:strongtemperature}
	\begin{gathered}
		\begin{cases}
			- \nabla \cdot \left(\right. \bm{\kappa} \nabla T \left.\right)  = Q  & \text{ in } \Omega , \\  T  = \bar{T}  & \text{ on } \mathcal{S}_{\text{D}}, \\  \left( \bm{\kappa}\nabla T\right)\cdot\bm{n}=\bar{q}   & \text{ on } \mathcal{S}_{\text{N}}, \\ \left( \bm{\kappa}\nabla T\right)\cdot\bm{n}=\bar{h}\left(T_{\infty}-T \right)    & \text{ on } \mathcal{S}_{\text{R}},
		\end{cases}
	\end{gathered}
\end{equation}
where $T$ represents temperature in the variable heat source problem and the hydraulic head in the Darcy flow problem, $\bm{\kappa}$ represents conductivity tensor, and $Q $ is source term. Let $v$ be test function that satisfies Dirichlet boundary condition, leading to the weak form of Eq. \ref{eq:strongtemperature} 

\begin{equation} \label{eq:weaktemperature}
	\begin{gathered}
		\int_{\Omega}\left[\nabla v \cdot \left( \bm{\kappa} \nabla T \right) -v Q\right] \mathrm{d} \Omega=\int_{\Omega}\left[ \nabla^\mathrm{T} v\bm{\kappa} \nabla T  -v Q\right] \mathrm{d} \Omega=0.
	\end{gathered}
\end{equation}	

When the $\bm{\kappa}$ is symmetric, the minimization form of Eq. \ref{eq:strongtemperature} exists

\begin{equation}\label{eq:temperaturefunctional}
	\begin{gathered}
		\min_T I=\frac{1}{2} \int_{\Omega}\left[ \nabla^\mathrm{T} T\bm{\kappa} \nabla T-2QT\right] \mathrm{d} \Omega,		
	\end{gathered}
\end{equation}

where $T$ satisfies Dirichlet boundary condition.

For variable heat source problem, the goal of VOL is to learn the mapping between heat source field and the temperature field. For Darcy flow problem, the goal of VOL is to learn the mapping between the conductivity field and the hydraulic head. For more details about problem settings of variable heat source problem and Darcy flow problem, see Supplementary material S2.1.

% \subsubsection{Indefinite Helmholtz Equation}

% Helmholtz equation has applications in many areas of physics, such as electromagnetic radiation, seismology, acoustics and quantum mechanics. Solving . We consider Helmholtz equation with Neumann boundary condition. Solving helmholtz equation numerically is difficult because 

% \begin{equation}
% 	\begin{gathered}
% 		\begin{cases}
% 		\left(-\Delta -\frac{\omega^2}{c^2} \right)u=0 &\text{in } \Omega \\
% 		\nabla u \cdot \bm{n} = h & \text{in } \mathcal{S}_\text{N},
% 		\end{cases}
% 	\end{gathered}
% \end{equation}

% Real case, the weak form is 

% \begin{equation}
% 	\begin{gathered}
% 		\int_{\Omega}\left(\nabla^\mathrm{T} v \cdot \nabla u  - v \frac{\omega^2}{c^2} u \right) \mathrm{d}  \Omega= \int_{\mathcal{S}} v \nabla u \cdot \bm{n}  \mathrm{d} \mathcal{S}
% 	\end{gathered}
% \end{equation}

\subsection{Scaling experiments}
\label{sec:sizescaling}

\begin{figure}[htbp]
	\centering
	\includegraphics[width=1\textwidth]{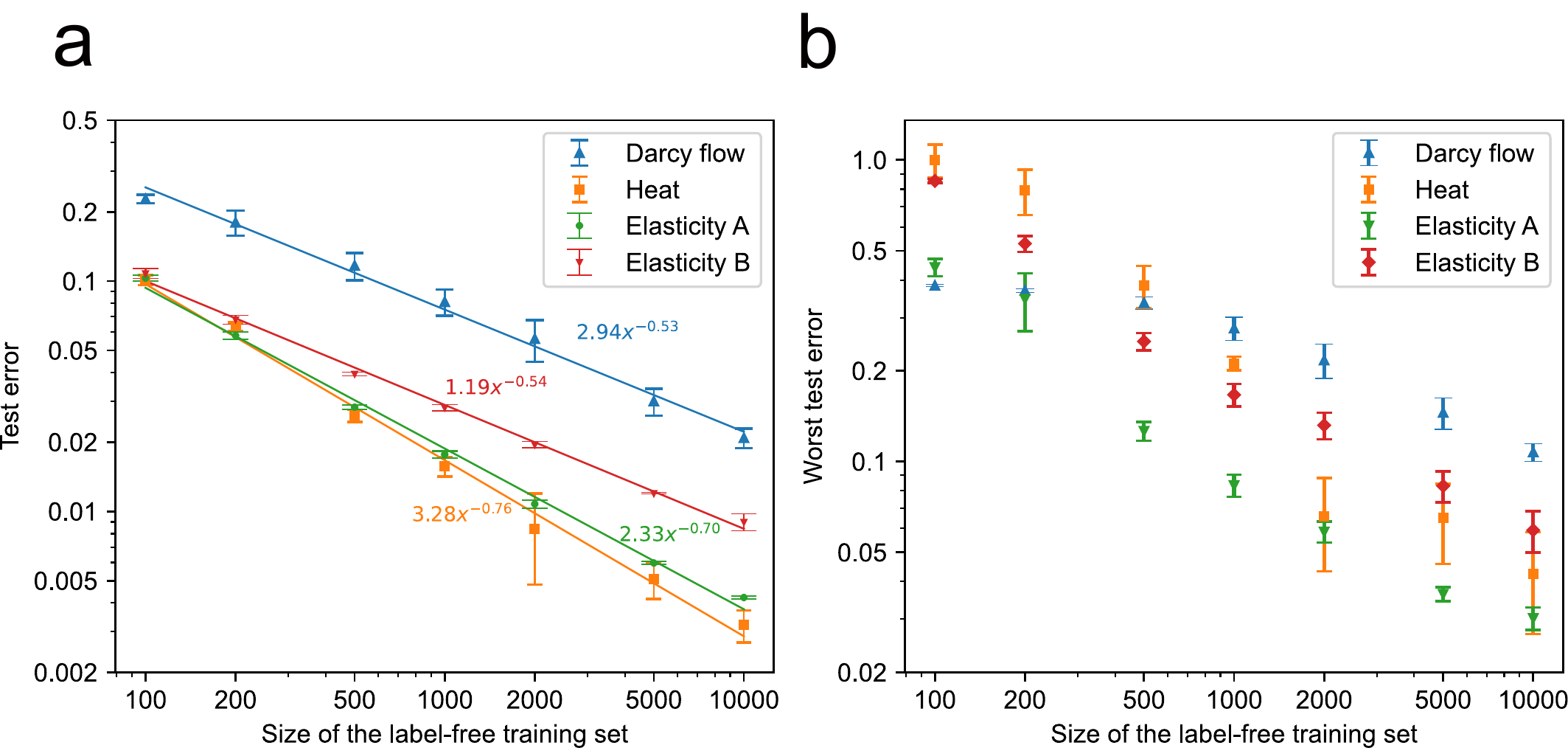}
	\caption{\textbf{Results of scaling experiments.} \textbf{a}, The test errors for all cases in the scaling experiments show a polynomial convergence rate. The error bars show the one standard deviation from 5 runs with different training/test data initialization. $x$ is the number of training data. \textbf{b}, The worst test errors also decrease in all cases as the size of training data increases.}
	\label{fig:scaling}
\end{figure}

% The error refers to the absolute error $\left\lvert \text{Prediction-Ground Truth}\right\rvert $.

To verify the effectiveness of VOL with different unlabeled data sizes, we design scaling experiments for all cases in section \ref{sec:problemsetting} with a specific resolution. For all cases, we chose 7 different sizes for the training set: 100, 200, 500, 1000, 2000, 5000, 10000, and set the size of the test set to 2000. All scaling experiments are conducted with VOL+CG(2) to demonstrate the capability and effectiveness of VOL to learn operators with small update steps. For more details about scaling experiments, see Supplementary material S6.1. With the increase of the number of unlabeled data for training, the performance of VOL should be improved considerably, which has been observed in Fig \ref{fig:scaling}a. With a log scale in data size and test error, an approximately linear convergence versus the size of training set is observed among all cases. From our size scaling results, the test error and the training set size matches a power law, which has also been observed in research about large language models (LLMs) \cite{kaplan_2020_scaling,maloney_2022_a} and DeepONet with the data-driven strategy \cite{Lu2021}, i.e., $y=ax^b$, where $y$ refers to the test error and $x$ refers to the training set size. Related coefficients have also been reported in Fig \ref{fig:scaling}a. In addition to the average metrics of VOL on the test set, we also study the worst-case scenario. Fig \ref{fig:scaling}b shows that the test errors of the worst prediction decrease for all benchmarks as the training set size increases. Extended Data Figure \ref{fig:worst12}, Extended Data Figure \ref{fig:worst3} and Extended Data Figure \ref{fig:worst4} visualize the worst predicted samples, which has largest test errors of all problems at ($N_{\text{training set}}=10000$, $N_{\text{test set}}=2000$), where we observe VOL still provides approximately correct solution fields.

% the test errors behave as a power law in the training set size
% With log 10 scale in x and y axis, we can observe the relationship between 
% We use the same training setting for all experiments in the same case. 
% And it is also unreal to try  
% Note the difficulty level of these cases cannot be simply determined by comparing their test errors from a single training setting (learning rate, weight deday, batch size, etc.).
% it is  does not necessarily indicate  them, 
\subsection{Resolution experiments}
\label{sec:resolutionscaling}

% Introduce the experiments. Explain the intuition. Analyse the results and Draw your conclusion.
% 
To demonstrate the capability of VOL at different resolutions, we conduct resolution experiments on two problems: the Darcy flow and elasticity B. We keep the size of the training set and the test set at all resolutions be 2000 for these two problems. For experiments at all resolutions in one problem, we use the same model architecture settings. For complete training settings and other details, see Supplementary material S6.2. For darcy problem, we observe the test error slightly rises as the resolution increases. For elasticity B problem, we observe the test error also slightly rises from resolution 33 to resolution 257$\times$257. With resolution 513$\times$513, we observe the test error drops to an average value of 1.89$\%$, we consider which is likely due to the adjustment of batch size at resolution 513$\times$513. Benefited from the resolution-invariant feature of FNOs \cite{li2020fourier}, VOL is able to learn solution operators mostly effectively at different resolutions with the same amount of parameters.

% Benefited from the resolution-invariant feature of FNOs \cite{li2020fourier}, VOL is able to learn solution different resolution with the same amount of parameters.
% is the same as the average test error at resolution 33. 
% Note we set batch size to $8$ for   
% We use the same training settings for all resolutions in one problem for comparison.
% For the $512\times512$ resolution, because the resolution-invariant feature of neural operator architecture, and 

\begin{table}[]
	\label{tab:resolutionexp}
	\begin{tabular}{ccccccc}
	\hline
	\multirow{2}{*}{Darcy flow}   & Resolution & 32 & 64 & 128 & 256 & 512 \\ \cline{2-7} 
								  & Test error & 1.14$\pm$0.04   & 1.25$\pm$0.07   &  1.41$\pm$0.08   &   1.69$\pm$0.18 &  2.21$\pm$0.32  \\ \hline
	\multirow{2}{*}{Elasticity B} & Resolution & 33 & 65 & 129 & 257 & 513 (batch size=8)   \\ \cline{2-7} 
								  & Test error & 1.89$\pm$0.03   & 1.92$\pm$0.04   &  1.94$\pm$0.04   & 1.95$\pm$0.06     & 1.89$\pm$0.02   \\ \hline
	\end{tabular}
	\caption{\textbf{Test errors of resolution experiments.} }
\end{table}
% darcy3264128
% 1.14$\pm$0.04
% 1.25$\pm$0.07
% 30.04$\pm$0.01
% 1.41$\pm$0.08
% \subsection{Investigation of optimization strategies}
% \label{sec:compareoptim}
% \subsection{Generalization benefits }
% \label{sec:generalizationbenefits}

\subsection{Generalization benefits}
\label{sec:generalizationbenefits}
% In this section, we investigate optimization strategies of VOL. We first discuss the effects of the number of update steps in VOL, and compare performance between VOL and the classical restart iterative method. Then, we compare iterative update with direct residual minimization and the classical data-driven method.
 
\begin{figure}[htbp]
	\centering
	\includegraphics[width=1\textwidth]{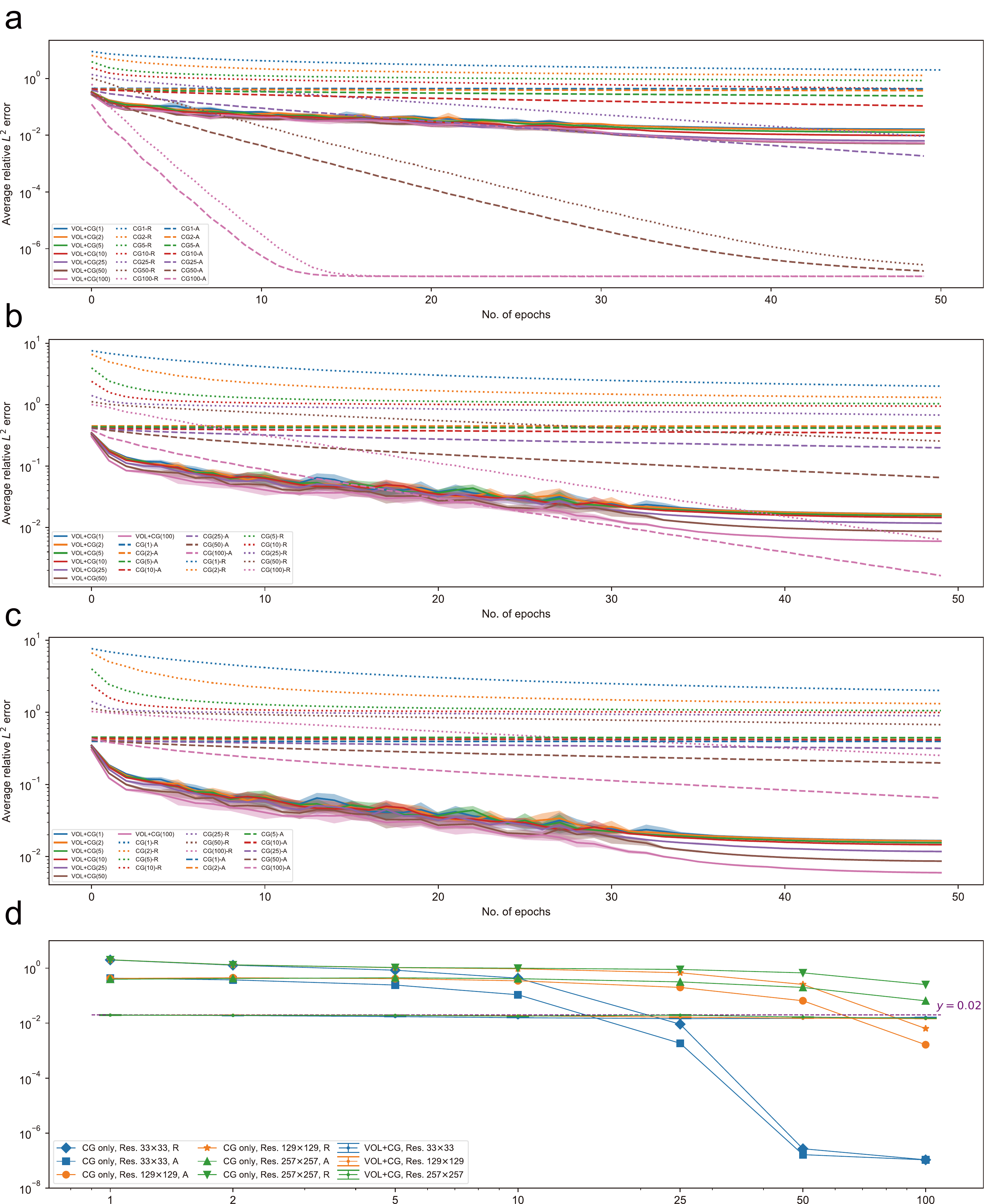}
	\caption{\textbf{Comparison on VOL and classical iterative methods.} \textbf{a}, \textbf{b}, \textbf{c}, Average relative $L^2$ errors on the training set are recorded for every epoch in the first experiment. \textbf{a}, \textbf{b}, \textbf{c} are the training errors at resolution 33$\times$33, 129$\times$129, 257$\times$257 respectively. The shaded regions denote one standard deviation. \textbf{d}, We plot $y=0.02$ as a reference line VOL+CG($i$). As the number of update steps increases, the test error of VOL+CG($i$) decreases slowly, from coinciding with $y=0.02$ to gradually falling below it.}
	\label{fig:comparingclassical}
\end{figure}

% , and training error refers to the relative $L^2$ error between the prediction and the label (See Supplementary materials section for more details)
% where the training error is set as average relative $L^2$ error on the training set per epoch
From the perspective of the neural solver, the \emph{generalization}, as a unique advantage of the machine learning, especially of the deep learning, can be utilized to provide better initial solutions for the solver solving new problems. In the context of VOL, generalization benefits means that the neural operator module keeps learning and provides more accurate initial solution for the iterative solver. To demonstrate the generalization benefits of VOL, we conduct two experiments, which compare VOL+CG($i$) and CG($i$) on same and different datasets respectively with various numbers of update steps on the elasticity B problem. Specifically, for a comprehensive and fair comparison, we choose restarted conjugate gradient CG($i$) with a set of numbers of update steps, $ i\in\left \{ 1, 2, 5, 10, 25, 50, 100 \right \} $, to compare with VOL that uses the same set of CG update step numbers for these two experiments (VOL+CG($i$)) in these two experiments. When $i=1$, conjugate gradient method degenerates itself to one-step steepest decent. We only study $\text{SD}=\text{CG}(1)$ case for these two experiments, and do not research steepest decent with more update steps, for SD has a well-known slow convergence speed, while more update steps of SD also brings more times of calculating matrix-vector products like CG. We run CG($i$) with its initial guess as the average of labels (CG($i$)-A) in the shift set used by VOL and with a random normal distribution initialization (CG($i$)-R) independently. For every epoch, CG($i$) solves each sample in the training set with $i$ update steps and restarts itself for the next epoch. We conduct these two experiments at resolution 33$\times$33, 129$\times$129 and 257$\times$257. See Supplementary material S6.3 for more details of the experimental design.

For the first experiment, we compare VOL and restarted conjugate gradient method on a same dataset containing 2000 samples, as shown in Fig. \ref{fig:comparingclassical}a to Fig. \ref{fig:comparingclassical}c. We set no test set for VOL in this experiment. At resolution 33$\times$33, when the number of update step is set to 1, 2, 5, 10, the train error of VOL is approximately one order of magnitude smaller than restarted CG at then end of iterations. As the number of update steps increases, restarted CG starts to converge faster. At resolution 33$\times$33, restarted CG has a faster convergence speed than VOL at 25, 50, 100 update steps, and has also a far smaller average train errors than VOL at 50 and 100 update steps at the end of training. At higher resolutions, however, classical iterative methods converge much slower. For resolution 257$\times$257, compared with CG($i$)-R, the training errors of VOL+CG($i$) are approximately one order to even two orders of magnitude smaller for all considered update steps at the end of iterations. Compared with CG($i$)-A, VOL+CG($i$) also has around one order of magnitude smaller training errors at the end of iterations. From Fig. \ref{fig:comparingclassical}a to Fig. \ref{fig:comparingclassical}c, we observe the error lines of CG($i$)-R and CG($i$)-A with the same update step are nearly parallel before their convergence, which indicates that, in our case, the difference of initialization has no significant effect on the convergence speed of the restart conjugate gradient method. We also note that generalization benefits, which can improve the magnitude of test errors, do not significantly improve the convergence speed of VOL+CG($i$). In fact, the condition number of coefficient matrix $\mathbf{K}$ is the key factor affecting convergence speed \cite{ysaad_2003_iterative}. We discuss possible solutions to accelerate the convergence as our future work (see section \ref{sec:conc}).

% In fact, the condition number of coefficient matrix $\mathbf{K}$ is the key factor affecting convergence speed \ref{}.  
% A better initialization does not necessarily accelerate solution,
% Advantage of VOL is evident from the end of the first epoch of training,

To verify the competitiveness of generalization for unseen data of VOL against the classical restarted iterative method, we conduct the second experiment, where the neural operators are first trained with VOL+CG($i$) on a 2000-sample training set, and then tested on a separate 2000-sample test set at the same resolution. Results (Fig. \ref{fig:comparingclassical}d) show that VOL can generalize to unseen data at various resolutions with different update step numbers, and the superiority of VOL becomes more and more obvious as the resolution grows. However, we also observe that VOL improves but has no significant improvement on the test error as the number of update steps grows, which needs further investigation. 

% train dataset containing 2000 samples at different resolutions, as shown in Fig. \ref{fig:experimentsgraph}
% The performance
% gain comes from the generalization of the neural operators. The neural operator keeps learning and provides
% more accurate initial solution for the solver, thus improving the performance.

% comparing VOL and classical iteration methods  with various numbers of update steps on the elasticity B problem
% shows our experiments ,

% brings a huge boost to the performance of the solvers  
% , the results of which show VOL brings a huge boost to the performance of the solvers.

% In this section, we compare the effects of SD and CG with different number of explicit and implicit Fourier layers. The dataset used in this section is formally introduced in section \ref{sec:vspcase2}. The common settings of these experiments in this section are shown in Table \ref{tab:compare_setting}.

\subsection{Comparison on different optimization strategies}
\label{sec:ddos}

\begin{figure}[htbp]
	\centering
	\includegraphics[width=1\textwidth]{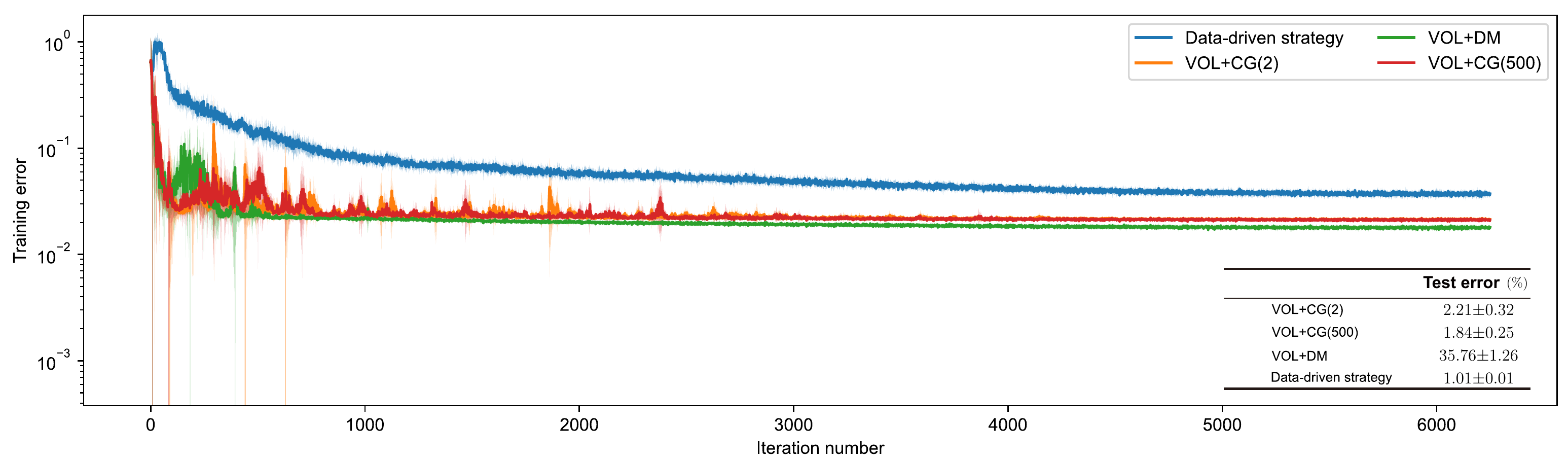}
	\caption{\textbf{Comparison on different optimization strategies.} For all strategies, the training error is set as average residual norm per batch, and is recorded for every iteration step. In this figure we also report the test errors (average relative $L^2$ error) at the lower right corner. }
	\label{fig:comparingpdd}
\end{figure}

In this section, we compare different optimization strategies for training neural operators with experimental results and discussion. We consider three strategies, iterative update (VOL+CG), direct minimization (VOL+DM), and purely data-driven strategy. For VOL+CG, we also chose 2 and 500 conjugate update steps respectively for VOL to further investigate the effect of update step number on performance of VOL. For all strategies, the experiment is conducted on Darcy flow problem at 512$\times$512 resolution, with 2000 training data and 2000 test data. To ensure a fair comparison, we keep all training settings (number of epochs, learning rate, scheduler, batch size, etc., for details see Supplementary material S6.4) same for each strategy. We observe no significant difference between VOL+CG(2) and VOL+CG(500) in training, but for test error VOL+CG(500) is a bit better than VOL+CG(2) on average. We then observe that while the converged residual norm of each strategy satisfies relationship "data-driven strategy $>$ VOL+CG $>$ VOL+DM", the relationship of test errors is the exact opposite of that, which is "data-driven strategy $<$ VOL+CG $<$ VOL+DM". For the data-driven strategy, the residual norm first rises to $\sim$10$^0$, and then decreases with a steep slope. As the training proceeds, the slope of the data-driven strategy steadily gets smaller util the end of training. For VOL+CG and VOL+DM, however, no rise of the residual norm was observed at start. In contrast, a sharp drop of the residual norm is observed from the very beginning of the training ($\sim$10 steps) for VOL+CG and VOL+DM. After the sharp drop, the residual norm of both VOL+CG and VOL+DM falls a bit in significant fluctuations. The distinguishable fluctuations of VOL+DM almost stop after $\sim$1000 iterations, while that of the VOL+CG stop after $\sim$3000 iterations. We also notice that though VOL can achieve a smaller residual norm than the data-driven strategy, the residual norm nearly stagnated after initial progress. In section \ref{sec:strategy}, we propose that VOL+CG can be interpreted as training neural operator module with provisional labels. Provisional labels, as intermediate solutions provided by the iterative solver, contain noise and are not accurate as ground truth labels, which might explain the gap of performance on test set between VOL and data-driven strategy. From experimental results, we also observe direct minimization has the worst test error in all strategies. Different from the strong form case \cite{Sirignano2018,raissi_2019_physicsinformed}, it might not be a good choice to penalize the residual by putting it in the loss function in our finite element settings. In fact, we can prove that VOL+DM strategy feeds the direct iterative update on the normal equation of the original linear system to the neural operator module (see Supplementary material S8 for proof), which might explain the gap of performance on test set between VOL+DM and other strategies. It is also noted that, even though comparative experiments show all strategies of VOL have larger average test errors than the data-driven strategy, the test errors of VOL+CG are just passable.

\section{Methods}
	\label{sec:VOL}
	% how to learn an operator.
	
\begin{figure}[htbp]
		\centering
		\includegraphics[width=1\textwidth]{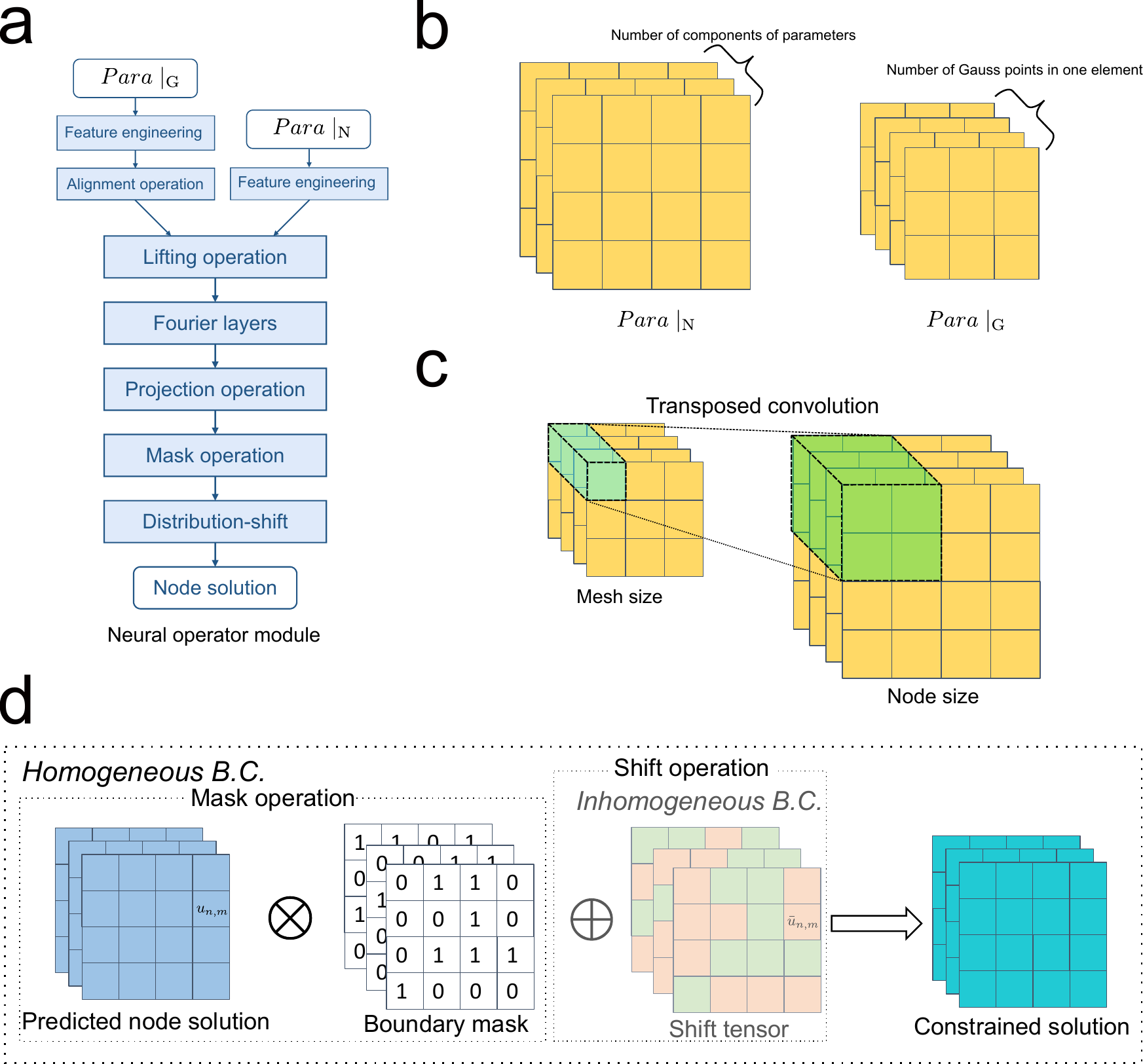}
		\caption{\textbf{Neural operator module and the related operations.} \textbf{a}, The neural operator module in the framework of VOL. \textbf{b}, Parameter tensor ($Para \mid_\text{G}$, left) aggregated by parameters discretized at Gauss points and parameter tensor ($Para \mid_\text{N}$, right) aggregated by parameters discretized at nodes. \textbf{c}, The alignment operation in the neural operator module. \textbf{d}, Mask operation and shift operation. {The homogeneous essential boundary conditions can be applied to the node solution by the mask operation. For shift operation, the inhomogeneous boundary condition can be then applied with a shift tensor, where the red elements denote the inhomogeneous terms, and the green elements denote zeros. For example, as shown in the figure, ${u}_{n,m}$ is the predicted solution to the $n$th degree of freedom of the $m$th node, the corresponding element of which in the mask tensor is 0. And the corresponding element in the shift tensor is $\bar{u}_{n,m}$, which denotes the inhomogeneous term of the constraint of $n$th degree of freedom of $m$th node.} 
		}
		\label{fig:inputparameters}
	\end{figure}

	%In this section, problem settings are first introduced in section \ref{sec:problemsetting}, and functional of the problem is constructed. Mode collapse phenomenon with total energy functional minimization is observed and investigated in sub section \ref{sec:modecollapse}. The failure of directly minimizing total energy strategy leads us to variational method, which is one of 
	%\subsection{Problem settings and functional approximation}
	%\label{sec:problemsettingandtpea}
	In this section, the variational operator learning (VOL) algorithm is developed. VOL does not focus on creating novel neural operator architectures, instead, it focuses on training the existing state-of-the-art neural operators with the smallest possible amount of labeled data, even no label. VOL is applicable to \emph{any field-wise neural architecture} in principle, but for the purpose of a clear demonstration of the basic effectiveness of VOL, in all experiments we just employ the Fourier neural operator (FNO) \cite{li2020fourier} in the neural operator module of VOL with moderate modification. Related network flowchart is shown in Fig. \ref{fig:inputparameters}a.
	
%	Generalization of deep learning makes the training (also solving) process efficient: while learning and solving a batch of samples, VOL can improve the potential performance on other samples. The overall process of VOL is shown in Fig. \ref{fig:architecture}. In section \ref{sec:problemsetting}, a general setting of elasticity is chosen, as the background to derive the principle of virtual work as a variational form of the governing equations. And the formulation of the functional of the continuum elastic system is given. In section \ref{sec:ritzapproach}, the approximation of functional used by VOL is derived, and the optimization objective of VOL is constructed.

		\begin{algorithm}[H]
		\begin{algorithmic}[1]
			\REQUIRE neural operator module $\mathcal{N}_{operator}\left( \bm{\vartheta  } \right) $, number of epochs $N$, optimization strategy $Opt$, learning rule $\eta$, and training set $\mathcal{D}$=$\left \{ \mathbf{U}_i \right \}_{i=1}^{N_\text{train}}$, max iteration number of one epoch $maxiter$, mask operation $Mask$, Shift set $Shift$, network optimizer $Opt_\text{net}$ and learning rate scheduler $Scheduler$
			\FOR{$1\leq i \leq N $} 
			\FOR{$1\leq j \leq maxiter $}
			\STATE Sample a minibatch of $bs$ examples $\mathbf{U}_{j}^{bs}$ from $\mathcal{D}$   
			\STATE $ \mathbf{a}_{j}^{bs}=\mathcal{N}_{operator}\left( \bm{\vartheta  } \right) \left( \mathbf{U}_{j}^{bs}, Mask, Shift\right) $
			\STATE Get $\mathbf{R}_j^{bs}$ with Ritz approach or Galerkin approach 
			\STATE $\mathbf{R}_j^{bs}= Mask\left( \mathbf{R}_j^{bs}\right) $ 		 
			\STATE $\Delta \mathbf{a}_{j}^{bs} = Opt\left( \mathbf{R}_j^{bs} , \mathbf{a}_{j}^{bs}, \mathbf{U}_{j}^{bs}, Mask\right)  $ 
			\STATE Get the current learning rate $\hat{\eta} $ according to $\eta$ and $Scheduler$
			\STATE Update $\bm{\vartheta }$ with $\Delta \mathbf{a}_{j}^{bs}$, $Opt_\text{net}$ and $\hat{\eta} $
			\ENDFOR
			\ENDFOR
			\ENSURE learned neural operator module $\mathcal{N}_{operator}\left( \bm{\vartheta  } \right) $
		\end{algorithmic}
		\caption{\textbf{Variational operator learning algorithm}}
		\label{alg:vol}
	\end{algorithm}
	
	The proposed VOL can be seamlessly integrated into deep learning training pipeline. The Algorithm \ref{alg:vol} demonstrates a complete training process including the outer epoch loop and the inner dataset loop. Here we just start discussion at the inner loop.
	
	Line 4 of the Algorithm \ref{alg:vol} shows a standard forward propagation from unlabeled input parameters to the node solution. Each channel of the node solution tensor contains one component of the node solution. Parameters of PDEs are first discretized at Gauss points and nodes, and then are aggregated into the parameter tensor $Para \mid_\text{G}$ and $Para \mid_\text{N}$ respectively (Fig. \ref{fig:inputparameters}b). For $Para \mid_\text{G}$, parameters at Gauss points that at the same position of all elements are encoded into one channel of the parameter tensor. Thus, the number of the channel of the parameter tensor equals to number of Gauss points in one element, and the resolution of $Para\mid_\text{G}$ is equal to the mesh size. For $Para \mid_\text{N}$, one channel of parameter tensor corresponds to one component of the parameter. For a concerned parameter of PDEs, the neural operator module (Fig. \ref{fig:inputparameters}a) receives $Para \mid_\text{G}$ or $Para \mid_\text{N}$ as input. Light feature engineering is first adopted for the input parameter tensor. The alignment operation (Fig. \ref{fig:inputparameters}c) is needed only for $Para \mid_\text{G}$ input, which is implemented with a trainable transposed convolution, mapping the tensor from the mesh size to the node size. Then, the lifting layer lifts the tensor to a higher channel space, $N$ Fourier layers are adopted, and the projection layer projects tensor to the solution space, which is the main operations of FNO \cite{li2020fourier}. Two additional operations are performed on the node solution predicted by the neural operator module:
	
	(1) Mask operation. The mask operation is designed to apply the essential boundary condition, such as displacement boundary condition in elasticity and temperature boundary condition in heat transfer to the system, which is equivalent to the constraint imposition process in the FEMs. First, a mask tensor that contains 0 and 1 is constructed, which has the same shape as the node solution. Every element in the mask tensor corresponds to a certain degree of freedom of a certain node in the computational mesh. If a element of the mask tensor is "0", it means that the corresponding degree of freedom of the corresponding node is constrained, while "1" means not constrained. Then, the element-wise product between the solution tensor and the mask tensor is calculated, as shown in Fig. \ref{fig:inputparameters}d. The technique of the mask operation can be categorized into the so-called \emph{hard} manners \cite{sun_2020_surrogate,Gao2021,rao_2023_encoding} to enforce the boundary condition of PDEs, which is different to the \emph{soft} manner, where the BCs are treated as penalty terms in the loss function. The mask operation can also be extended to the inhomogeneous case. As shown in Fig. \ref{fig:inputparameters}d, we simply add a shift tensor with the same shape after the element-wise product operation, the elements of which are 0 where the corresponding elements of the mask tensor are 1 (unconstrained), and the other elements are the inhomogeneous terms of constraints. 
	
	(2) Distribution-shift. Distribution-shift operation that exists in the prior work \cite{li2020fourier,you_2022_learning} is remained in this work, but we only use a very small number labeled data (5 labels) rather than label the whole training set. It first computes the mean $mean$ and the standard deviation $std$ of all labels of the training set, and then use the following equation to shift the output of the neural operators to the distribution of the labels:
	\begin{equation}\label{eq:distributionshift}
		\begin{gathered}
			output = output\otimes std+ mean
		\end{gathered}
	\end{equation}
	
	Such a shift operation can stabilize the training process. Note the training set is totally label-free. To generate labels for the shift operation, an extra small batch of parameters of PDEs are randomly sampled and labeled, and these labeled data are only used in the distribution-shift operation, which we call a \emph{shift set}.
	
	\subsection{Matrix-free approximation of the system functional and the system residual}\label{sec:matrixfree}
	
	In line 5 of Algorithm \ref{alg:vol}, Ritz approach and Galerkin approach of VOL are developed respectively to approximate the system functional and the system residual. In this section, we introduce these two approaches with theory consideration and implementation details.
	
	% Ritz approach first approximates the system functional, and then utilizes automatic differentiation to get the residual, while Galerkin approach approximates the residual with the test kernel. Galerkin approach is backpropagation-free.
	
	\subsubsection{Ritz approach}\label{sec:ritzapproach}

	%In section \ref{sec:problemsetting}, functional of elastic continuum is given in Eq. \ref{eq:continuumtotal}.
	Ritz approach of VOL is composed of two parts, system functional approximation (forward propagation) and residual calculation (backpropagation). Ritz approach is matrix-free and can approximate system functional and system residual numerically. In this section, an approximation of the system functional used by Ritz approach is first introduced with the context of elasticity, the formulation of which is a bit different from that in FEMs, making VOL possible to approximate system functional without calculating stiffness matrices. Then, we introduce how to calculate system functional numerically with deep learning toolkit and the residual calculation. It is worth noting that it is possible for VOL to use various discretization scheme, notwithstanding VOL uses the same discretization scheme (piecewise polynomial interpolation) as FEMs in this Article:
	\begin{equation}\label{eq:piecewise polynomial interpolation}
		\begin{gathered}
			\mathbf{u}\left( x,y,z\right) =\mathbf{N}\left(r,s,t \right)  \mathbf{a}^e,\\
			\mathbf{x}=\mathbf{N}\left(r,s,t \right)\mathbf{x}^e.
		\end{gathered}
	\end{equation}
	%The VOL takes the same discretization scheme as FEMs in this Article
	%We also compare the difference between introduce the variational operation of VOL.  
	%VOL approximates functional and conducts variational online zhejuhua fangdao zuihou? 
	%In this Article, VOL takes the same discreti..scheme zhegefangzaizuihoushuo
	% in structured meshing zai suanli bufen shuo
	%By piecewise polynomial interpolation, 
	Note $\mathbf{u}\left( x,y,z\right)$ and $\mathbf{a}^{e}$ are node-related quantities, they can represent the displacement in the context of solid mechanics, and the temperature in heat transfer, etc.
	
	We take an elastic body discretized by isoparametric elements as an example. In this case, the node solution is the node displacement tensor of the elastic body. The classical expression of functional approximation in FEMs is:
	%We take an thin elastic plate discretized by quadrilateral isoparametric elements with four nodes in structured meshing as an example. In this case, the node solution are the node displacement of the elastic plate. The classical expression of functional approximation in FEMs is:
	%We take an elastic plate as one example, and transfer the discretization scheme in FEMs to discretize the original problem, and implement it in tensor style. 
	%the quadrilateral element
	%ile suanfanenggouchuli 
	%caiyonglehe FEM xiangtongde formulation but bujuxianyu FEM 
	
	\begin{equation}\label{eq:discretetotal}
		\begin{aligned}
			\widetilde{\Pi}= & \sum_e \Pi^e=\frac{1}{2}\sum_e\left(\mathbf{a}^{e \mathrm{T}} \int_{\Omega_e} \mathbf{B}^{\mathrm{T}} \mathbf{D B} \mathrm{d} \Omega \mathbf{a}^e\right) 
			\\  
			&-\sum_e\left(\mathbf{a}^{e \mathrm{T}} \int_{\Omega_e} \mathbf{N}^{\mathrm{T}} \mathbf{f} \mathrm{d} \Omega\right)-\sum_e\left(\mathbf{a}^{e \mathrm{T}}\int_{\mathcal{S}_\sigma^e} \mathbf{N}^{\mathrm{T}} \mathbf{\bar{X}} \mathrm{d } \mathcal{S} \right).
		\end{aligned}
	\end{equation}
	The $\int_{\Omega_e} \mathbf{B}^{\mathrm{T}} \mathbf{D B} \mathrm{d} \Omega$, $\int_{\Omega_e} \mathbf{N}^{\mathrm{T}} \mathbf{f} \mathrm{d} \Omega$ and $\int_{\mathcal{S}_\sigma^e} \mathbf{N}^{\mathrm{T}} \mathbf{\bar{X}} \mathrm{d } \mathcal{S}$ in Eq. \ref{eq:discretetotal} are element stiffness matrix, element volume load vector and element face load vector respectively. Denote them as
	\begin{equation}\label{eq:noteele}
		\begin{gathered}
			\mathbf{K}^e=\int_{\Omega_e} \mathbf{B}^{\mathrm{T}} \mathbf{D B} \mathrm{d} \Omega, \quad
			\mathbf{P}^e=\int_{\Omega_e} \mathbf{N}^{\mathrm{T}} \mathbf{f} \mathrm{d} \Omega+\int_{\mathcal{S}_\sigma^e} \mathbf{N}^{\mathrm{T}} \mathbf{\bar{X}} \mathrm{d } \mathcal{S}.
		\end{gathered}
	\end{equation}
	We have
	\begin{equation}\label{eq:femtotal}
		\begin{gathered}
			\widetilde{\Pi}=\frac{1}{2}  \mathbf{a}^{\mathrm{T}}  \sum_{e}\left(\mathbf{G}^{e \mathrm{T}} \mathbf{K}^e \mathbf{G}^{e}\right) \mathbf{a}-\mathbf{a}^{\mathrm{T}} \sum_{e}\left(\mathbf{G}^{e \mathrm{T}} \mathbf{P}^{\mathrm{e}}\right),
		\end{gathered}
	\end{equation}
	and
	\begin{equation}\label{eq:globalKandP}
		\begin{gathered}
			\mathbf{K}=\sum_{e} \mathbf{G}^{e \mathrm{T}} \mathbf{K}^{e}\mathbf{G}^{e} ,\quad \mathbf{P}=\sum_{e} \mathbf{G}^{e \mathrm{T}} \mathbf{P}^e.
		\end{gathered}
	\end{equation}

	In FEMs, element stiffness matrices in Eq. \ref{eq:noteele} are first obtained by numerical integration and then are assembled together with Eq. \ref{eq:globalKandP} to form the global stiffness matrix. On the other hand, the whole process of Ritz approach is \emph{matrix-free}, i.e., it does not calculate any stiffness matrices or restore them, which is elaborated in the following paragraphs. The formulation of element stiffness matrix given by Gaussian quadrature method is:
	% which is elaborated in section \ref{sec:methodimplemention}
	\begin{equation}\label{eq:numerialKe}
		\mathbf{K}^e=\int_{\Omega_e} \mathbf{B}^{\mathrm{T}} \mathbf{D B} \mathrm{d} \Omega \approx \sum_{l=1}^{n_g} H_l \mathbf{B}_l^{\mathrm{T}} \mathbf{D}_l \mathbf{B}_l \left | \mathbf{J}_l^e\right |. 
	\end{equation}
	Similarly, 
	\begin{equation}\label{eq:numerialPe}
		\mathbf{P}^e \approx \sum_{l=1}^{n_g} H_l \mathbf{N}_l^{\mathrm{T}} \mathbf{f}_l \left | \mathbf{J}_l^e\right | +  \sum_{m=1}^{n_{\mathcal{S}_\sigma^e}} \sum_{l=1}^{n_m} I_{ml}\mathbf{N}_{ml}^{\mathrm{T}} \mathbf{\bar{X}}_{ml} \left | \mathbf{J}_{ml}^{\mathcal{S}_\sigma^e}\right |.
	\end{equation}

Substitute Eq. \ref{eq:numerialKe}, Eq. \ref{eq:numerialPe} into Eq. \ref{eq:discretetotal}, and note
	\begin{equation}\label{eq:strain}
		\begin{gathered}
			\bm{\varepsilon}=\mathbf{B} \mathbf{a}^e.
		\end{gathered}
	\end{equation}
	Then we have another kind of formulation to express the functional, which is slightly different from Eq. \ref{eq:discretetotal} and Eq. \ref{eq:femtotal}:
	\begin{equation}\label{eq:discretetotal2}
		\begin{aligned}
			\widetilde{\Pi}= & \sum_e \Pi^e \approx \frac{1}{2} \sum_e \sum_{l=1}^{n_g} H_l \mathbf{a}^{e \mathrm{T}} \mathbf{B}_l^{\mathrm{T}} \mathbf{D}_l \mathbf{B}_l \mathbf{a}^{e}  \left | \mathbf{J}_l^e\right |  
			\\  
			&-\sum_e \sum_{l=1}^{n_g} H_l \mathbf{a}^{e \mathrm{T}} \mathbf{N}_l^{\mathrm{T}} \mathbf{f}_l \left | \mathbf{J}_l^e\right |-\sum_e \sum_{m=1}^{n_{\mathcal{S}_\sigma^e}} \sum_{l=1}^{n_m} I_{ml} \mathbf{a}^{e \mathrm{T}} \mathbf{N}_{ml}^{\mathrm{T}} \mathbf{\bar{X}}_{ml} \left | \mathbf{J}_{ml}^{\mathcal{S}_\sigma^e}\right |		\\
			or \\
			\widetilde{\Pi}= & \sum_e \Pi^e \approx \frac{1}{2} \sum_e \sum_{l=1}^{n_g} H_l \bm{\varepsilon}_l^{\mathrm{T}} \mathbf{D}_l \bm{\varepsilon}_l \left | \mathbf{J}_l^e\right |  
			\\  
			&-\sum_e \sum_{l=1}^{n_g} H_l \mathbf{a}^{e \mathrm{T}} \mathbf{N}_l^{\mathrm{T}} \mathbf{f}_l \left | \mathbf{J}_l^e\right |-\sum_e \sum_{m=1}^{n_{\mathcal{S}_\sigma^e}} \sum_{l=1}^{n_m} I_{ml} \mathbf{a}^{e \mathrm{T}} \mathbf{N}_{ml}^{\mathrm{T}} \mathbf{\bar{X}}_{ml} \left | \mathbf{J}_{ml}^{\mathcal{S}_\sigma^e}\right |
		\end{aligned}
	\end{equation}
	%FEMs uses formulations like Eq. \ref{eq:discretetotal1} to approximate functional, while VOLs 
	%The difference  Eq. \ref{eq:femtotal} and Eq. \ref{eq:discretetotal} and Eq. \ref{eq:femtotal}
	Ritz approach uses formulations like Eq. \ref{eq:discretetotal2} to approximate functional, while FEMs use Eq. \ref{eq:discretetotal} or Eq. \ref{eq:femtotal}. It is clear that neither the element stiffness matrix nor the global stiffness matrix appears in Eq. \ref{eq:discretetotal2}. Thus, to calculate Eq. \ref{eq:discretetotal2}, the calculation and the assembly process of the element stiffness matrices can be simply skipped.
	
	%According to Eq. \ref{eq:VP}, the solution field 
	According to Eq. \ref{eq:VP} and Eq. \ref{eq:piecewise polynomial interpolation}, the node solution field should satisfy
	\begin{equation}\label{eq:zhankai}
		\begin{gathered}
			\delta \widetilde{\Pi}= \frac{\partial \widetilde{\Pi}}{\partial a_1} \delta a_1 + 
			\frac{\partial \widetilde{\Pi}}{\partial a_2} \delta a_2+ \cdots + \frac{\partial \widetilde{\Pi}}{\partial a_{n_f}} \delta a_{n_f}=0
		\end{gathered}
	\end{equation}
	Eq. \ref{eq:zhankai} is also called the \emph{stationary condition} of $\widetilde{\Pi}$. Since $\delta a_1, \delta a_2, \cdots, \delta a_{n_f}$ are the virtual displacement of nodes, we have
	\begin{equation}\label{eq:variational}
		\begin{gathered}
			\frac{\partial \widetilde{\Pi}}{\partial \mathbf{a}}=\mathbf{0}
		\end{gathered}
	\end{equation}
	Substituting Eq. \ref{eq:femtotal} and Eq. \ref{eq:globalKandP} into Eq. \ref{eq:variational}, we have
	\begin{equation}\label{eq:KadengyuP}
		\begin{gathered}
			\mathbf{K}\mathbf{a}=\mathbf{P}
		\end{gathered}
	\end{equation}
	
	FEMs approximate the functional (Eq. \ref{eq:femtotal}), perform variational operation (Eq. \ref{eq:variational}) \emph{offline}, form and solve Eq. \ref{eq:KadengyuP} \emph{online}. On the other hand, Ritz approach of approximates the functional (Eq. \ref{eq:discretetotal2}) and performs the variational operation \emph{online} after every forward propagation. Ritz approach of VOL performs the variational operation to get the residual $\mathbf{R}$ of the predicted node solution field:
	\begin{equation}\label{eq:residualerror}
		\begin{gathered}
			\mathbf{R}=\frac{\partial \widetilde{\Pi}}{\partial \mathbf{a}}=\mathbf{K}\mathbf{a}-\mathbf{P}
		\end{gathered}
	\end{equation}
	% keyixingchengyizhangbiao table to compare FEM and VOL

	The deep Ritz method \cite{Weinan2018} and the literature \cite{samaniego_2020_an} also construct the system functional, but they just simply set functional minimization as the goal of optimization (minimization form), that is, they treat the system functional as the loss function in deep learning pipeline. Compared to the idea of deep Ritz method, our approach goes a step further. Rather than minimize the functional directly, we choose to set minimizing the norm of residual that derived from the variational operation as the optimization objective. Apart from a basic strategy of setting the norm of the residual itself as the loss function, we also turn to solve the corresponding linear system with iterative methods, by utilizing the connection between the computational mesh (see Extended Data Figure \ref{fig:residual--equation}), the residual and the linear system to minimize the norm of the residual. Such a scheme allows us to train on a mesh. (see section \ref{sec:strategy} for details of optimization strategy of VOL) 
	
	% We also tried minimization form directly like what deep Ritz method does, but result seems not satisfactory (for more information, see Supplementary materials section ).  

Four steps are needed to conduct numerical integration and get the system functional approximation of Eq. \ref{eq:discretetotal2} in the forward propagation of Ritz approach: 

(1) Calculate the weighted sum of $\mathbf{a}^e$ and other types of node-related physical quantities (such as discrete parameter field at nodes) weighted by the interpolation function and its spatial derivatives at Gauss points. Here we introduce the weighted sum of node solution. The same operations will be performed on other types of node-related physical quantities if needed in VOL. Observe Eq. \ref{eq:piecewise polynomial interpolation}, the left term of which can be seen as the weighted sum of $\mathbf{a}^e$:
\begin{equation}\label{eq:weighted sum}
	\begin{gathered}
		u_i = \sum_{j=1}^{M} N_j a^e_{j,i},
	\end{gathered}
\end{equation}
and take the derivative of both sides 
\begin{equation}\label{eq:weighted sum 2}
	\begin{gathered}
		\frac{\partial u_i}{\partial \mathbf{x}} = \sum_{j=1}^{M} \mathbf{J}^{e-1} \frac{\partial N_j}{\partial \mathbf{r}} a^e_{j,i}.\\
	\end{gathered}
\end{equation}
 
Eq. \ref{eq:weighted sum} and Eq. \ref{eq:weighted sum 2} are essentially equivalent to convolution operations on discrete node solution field, if the weight of the convolution filters is set as values of terms $N_j$ and $\mathbf{J}^{e-1} \frac{\partial N_j}{\partial \mathbf{r}}$ at Gauss points. On this basis, we develop the method of calculating Eq. \ref{eq:weighted sum} and Eq. \ref{eq:weighted sum 2} based on the standard non-trainable convolution operation, which is implemented easily with tensor-based deep learning engines:
\begin{itemize}
    \item[\ding{172}] Compute the value of $N_j$ and $\mathbf{J}^{e-1} \frac{\partial N_j}{\partial \mathbf{r}}$ at Gauss points, organize and restore them as convolution filters, which we name a trial kernel $\mathcal{K}_{\text{trial}}$. The trial kernel reflects the ansatz of the trial function. 
    \item[\ding{173}] Convolve $\mathbf{a}$ with the $\mathcal{K}_{\text{trial}}$, unit stride and no padding to get the feature map at Gauss points: $[u_i\mid_{\text{G}}, \frac{\partial u_i}{\partial \mathbf{x}}\mid_{\text{G}}]=\mathcal{K}_{\text{trial}} \circledast \mathbf{a}$. 
    
	% , \text{where} \circledast \text{has unit stride and no padding}

\end{itemize} 

An illustrative example of weighted sum convolution and calculating $\mathcal{K}_{\text{trial}}$ is given in Fig. \ref{fig:ritzandgalerkin}a and Fig. \ref{fig:ritzandgalerkin}d respectively.

% As a simplified implementation of discretization scheme, we use convolution to calculate weighted sum, which can not only be applied at Gauss points, but also can be applied anywhere else of the element domain in principle. Such way of weighted sum with convolution is well suited for the usage of isoparametric elements, that allow the elements that do not maintain orthogonality between the sides of the element. Such techniques provide pre-computed convolution kernels, which play a similar role to the stencils in FDM.

(2) Calculate the integrand.
In step (1), we get two types of weighted sum at Gauss points of all elements, i.e., $u_i$ and $\frac{\partial u_i}{\partial x_j} $, that gathered as tensors. In this step, as shown in Fig. \ref{fig:ritzandgalerkin}b, these weighted sum tensors are first reorganized into feature maps $\mathcal{F}_{\text{physics}}$ of physical properties in problems of concern (see section \ref{sec:elasticproblemsetting}) to facilitate the calculation of integrand with Eq. \ref{eq:discretetotal2} and the following Galerkin approach. Then, the integrand is then formed by calculating Eq. \ref{eq:discretetotal2} with $\mathcal{F}_{\text{physics}}$. The process utilizes the idea of \emph{domain knowledge embedding} in the section \ref{sec:relatedwork}. Note that this step varies in implementation details for governing equations in different domains.

(3) Multiply Jacobian (Fig. \ref{fig:ritzandgalerkin}c). The Jacobian tensor that contains the values of the Jacobian at Gauss points of all elements is pre-computed, according to the configuration of all elements in the physical space. Then, the element-wise product of the integrand tensor and the Jacobian tensor are computed to obtain the integrand tensor that considers Jacobian effect.

(4) Calculate the functional approximation of the system with numerical integration (Fig. \ref{fig:ritzandgalerkin}c). First, perform Gaussian quadrature for all elements, i.e., calculate weighted sum of the integrand by weight of Gauss points in every element. The element of result tensor is the functional approximation of the element $\Pi^e$. Then, sum up $\Pi^e$ of all elements to get the functional approximation $\widetilde{\Pi}$.

Since the forward propagation from node solution to the functional has been constructed, the gradient of the functional to the node solution can be easily derived with automatic differentiation. A backward propagation from the functional to the node solution is conducted, to obtain the gradient of the functional with respect to the node solution. The gradient of the functional to the node solution is just the residual of the linear system, so it is denoted as the residual tensor $\mathbf{R}$.

As discussed above,a forward-backward propagation loop between the node solution and the functional approximation has been constructed. The developed loop allows us to derive the residual of the linear system without acquiring element stiffness matrices and assembling the global stiffness matrix. The residual tensor $\mathbf{R}=\frac{\partial \widetilde{\Pi}}{\partial \mathbf{a}}$ is derived by just running the loop once, which has the same shape as the node solution.

	\subsubsection{Galerkin approach}\label{sec:galerkinapproach}

	In this section, Galerkin approach is introduced. Galerkin approach is another matrix-free method in VOL of calculating matrix-vector product. Compared to Ritz approach, Galerkin approach is much simpler and applicable even when the minimization form of PDEs does not exist. In contrast to Ritz approach, which follows $\mathbf{a} \to \widetilde{\Pi} \to \mathbf{R} $ procedure, Galerkin approach can derive $\mathbf{R}$ without calculating $\widetilde{\Pi}$ and backpropagation.

\begin{figure}[htbp]
		\centering
		\includegraphics[width=1\textwidth]{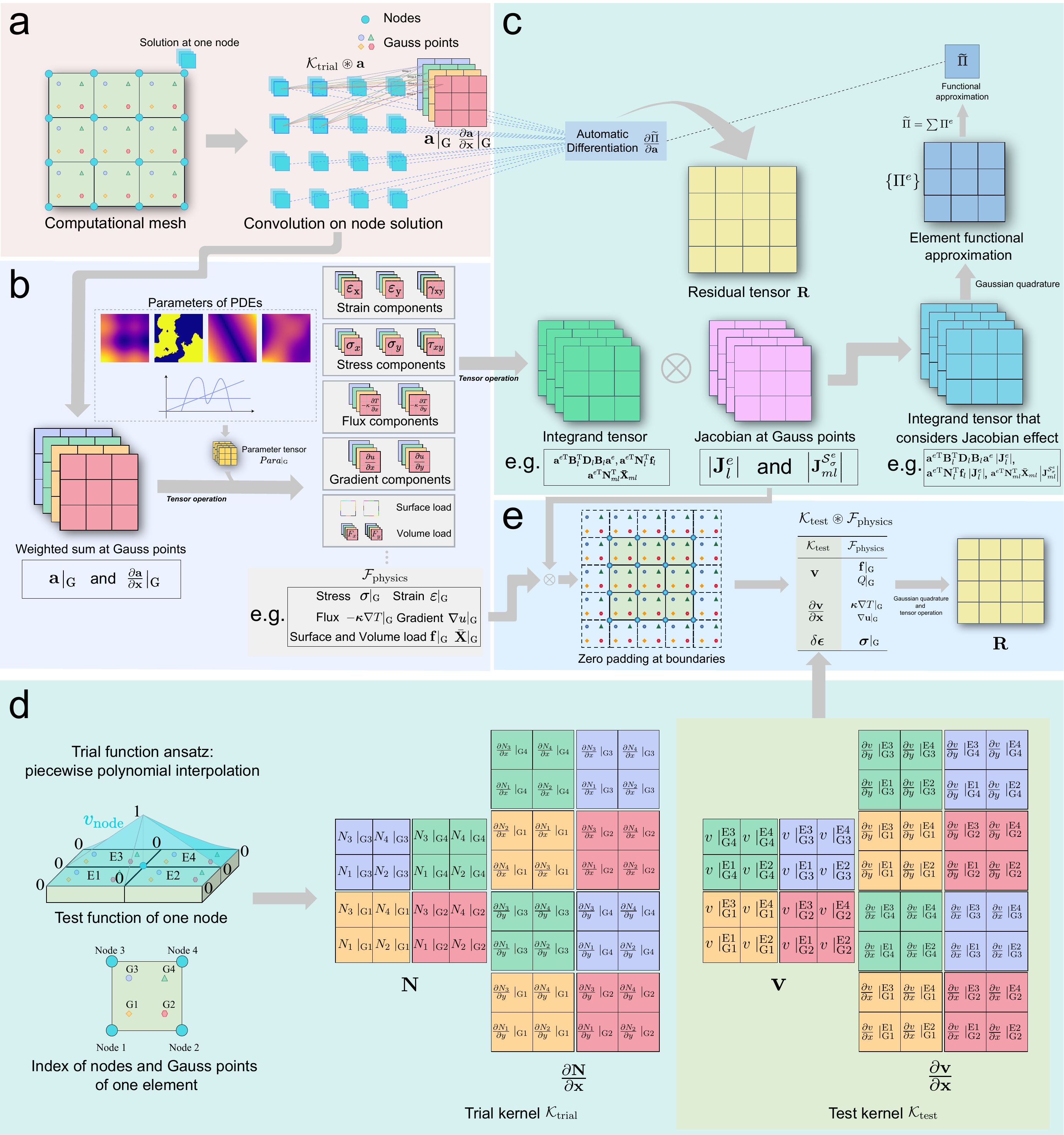}
		\caption{\textbf{Ritz approach and Galerkin approach of VOL.} In this figure, we introduce Ritz and Galerkin approach on a domain discretized by 9 quadrilateral elements with 4 nodes and 4 Gauss points as an example. \textbf{a}, Evaluation of the weighted sum of node solution at Gauss points with the convolution. Note node solution may have more than one component, and convolution operation on one component is plotted here for simplicity. In \textbf{a}, the width$\times$height of one filter is 2$\times$2, and 4 filters are needed. The shape of a weighted sum tensor is 4$\times$3$\times$3. \textbf{b}, Calculation of feature maps of physical properties $\mathcal{F}_{\text{physics}}$. Parameter tensor, and From $\mathbf{a}$ and $\frac{\partial \mathbf{a}}{\partial \mathbf{x}}$ and and a feature map of a component of physical properties \textbf{c}, System functional approximation and automatic differentiation in Ritz approach. Integrand tensor, Jacobian both have a shape of 4$\times$3$\times$3. $\left \{ \Pi^e \right \} $ has a shape of 1$\times$3$\times$3. \textbf{d}, The construction of $\mathcal{K}_{\text{trial}}$ and $\mathcal{K}_{\text{test}}$. \textbf{e}, The convolution between $\mathcal{K}_{\text{test}}$ and $\mathcal{F}_{\text{physics}}$ to derive $\mathbf{R}$ in Galerkin approach.}
		\label{fig:ritzandgalerkin}
	\end{figure}

	Like Ritz approach, Galerkin approach also needs $\mathcal{F}_{\text{physics}}$. So it also shares exactly the same convolution operation and kernel $\mathcal{K}_{\text{trial}}$ as Ritz approach to calculate $\mathbf{a}\mid_{\text{G}}$ and $\frac{\partial \mathbf{a}}{\partial \mathbf{x}} \mid_{\text{G}}$. Following classical Galerkin method, Galerkin approach first constructs the test function $\mathbf{v}$ in the deep learning engine. Here we simply consider $\mathbf{v}$ as the basis of $V_n$. Observing $\mathbf{v}$ of one node in Fig. \ref{fig:ritzandgalerkin}d, which has compact support and is non-zero over the adjacent elements of the node, we find it possible to arrange value $\mathbf{v}$ and $\frac{\partial \mathbf{v}}{\partial \mathbf{x}}$ at Gauss points into groups of filters according to the following rules:
	\begin{enumerate}[label=(\arabic*)]
		\item For each component of $\mathbf{v}$ and $\frac{\partial \mathbf{v}}{\partial \mathbf{x}}$, the value at the same Gauss point of all adjacent elements of the node is put into a separate filter.
		\item Value from different elements is arranged in the same order for all filters. 
	\end{enumerate}
	We name the filters as the test kernel $\mathcal{K}_{\text{test}}$, a 2-dim example of which is given in \ref{fig:ritzandgalerkin}d. Sometimes we may need to construct more sophisticated filters than $\mathbf{v}$ and $\frac{\partial \mathbf{v}}{\partial \mathbf{x}}$. For example, in elasticity we need to construct virtual strain as a test kernel $\mathcal{K}_{\text{test}}=\delta\bm{\varepsilon}=\frac{1}{2}( v_{i,j}+v_{j,i}) $. In this regard, we just calculate the components of the desired virtual quantity. 
	
	After constructing $\mathcal{K}_{\text{test}}$, we can express the weak form of Eq. \ref{eq:weakabstractform} with a single standard convolution operation on $\mathcal{F}_{\text{physics}}$ with the kernel $\mathcal{K}_{\text{test}}$ (Fig. \ref{fig:ritzandgalerkin}e). We also need to use the Jacobian tensor to scale $\mathcal{F}_{\text{physics}}$ before the convolution. To ensure that our operation is valid for nodes on the boundary, we need to add a zero padding to $\mathcal{F}_{\text{physics}}$, which represent a layer of elements wrapped around the solution domain with their $\mathcal{F}_{\text{physics}}$ set to 0. The convolution stride is set to 1. 

	% the $\mathcal{F}_{\text{physics}}$ of 
	% For convolution between $\mathcal{K}_{\text{test}}$ and $\mathcal{F}_{\text{physics}}$, 
	% $\mathcal{K}_{\text{test}}=\delta \bm{\varepsilon}=\frac{1}{2}( \delta u_{i,j}+ \delta u_{j,i}) $
	% Write discrete form of Eq. \ref{eq:weakform},
	% In context of group convolution, the group size of the filters is 1. 
	
	% To manipulate feature map of physical properties

	% has two main steps: 
	% \begin{enumerate}
	% 	\item  share the same step with Ritz approach$\mathbf{R} = \mathbf{A} \mathbf{a}$
	% 	\item Calculate the residual
	% \end{enumerate}
	In VOL, we use Ritz approach to calculate quadratic forms in SD and CG iterations. System residual and other matrix-vector product (like $\mathbf{Kp}$ in Algorithm \ref{alg:steepestdecent} and \ref{alg:CG}) are mainly calculated with Galerkin approach. We do not use Ritz approach to calculate system residual because doing so would bring additional backpropagation operations and cannot be extended to more general cases where the minimization form of PDEs do not exist.
	
	% Faster than Ritz approach. Reduces memory consumption. Autograd is memory-consuming. We only need one backpropagation. Compared to PINN etal, we do not need high order derivatives. The kernel of the convolution is the test function. Test function is applied to the convolution kernel.
	
	\subsection{Optimization strategy}
	\label{sec:strategy}
	
	The optimization objective of VOL is to find parameters of the neural operator that minimize the average of residual norm of the whole training set:
	\begin{equation}\label{eq:goal}
		\begin{gathered}
			\mathop{\min}\limits_{\mathbf{a}_i} \frac{1}{D} \sum \vert\vert \mathbf{R}_i \vert\vert_{i=1}^{D} \\
			\text{s.t. } \mathbf{R}_i=\mathbf{K}\mathbf{a}_i-\mathbf{P}
		\end{gathered}
	\end{equation}
	
	For implementation, the residual $\mathbf{R}$ is treated as a tensor. Thus, we just calculate the norm of the flattened $\mathbf{R}$ as the residual norm. 	The element of the residual tensor is the just the residual of corresponding equation of the linear system Eq. \ref{eq:KadengyuP}, the same one that solved by FEMs, as shown in Extended Data Fig. \ref{fig:residual--equation}. Thus, the goal of VOL (Eq. \ref{eq:goal}) has turned to solve all linear systems derived from the parameters of PDEs in the training set. Note that the residual tensor $\mathbf{R}$ now contains all residuals of the linear system. We need to zero out those elements in $\mathbf{R}$ that correspond to the constrained part of $\mathbf{a}$, so that the coefficient matrix of the linear system is non-singular. Thus, an additional mask operation (Fig. \ref{fig:inputparameters}) needs to be imposed on the residual tensor $\mathbf{R}$ in line 6.
	
	% \begin{figure}[htbp]
	% 	\centering
	% 	\includegraphics[width=1\textwidth]{architecture.pdf}
	% 	\caption{\textbf{The framework of variational operator learning (VOL).} The main operations, modules, and a forward and a backward propagation process of VOL are illustrated.}
	% 	\label{fig:architecture}
	% \end{figure}
	%\subsection{Variational operation implementation utilizing automatic differentiation}
	%\label{sec:methodimplemention}
	% bianfengongshi both fem and vol douyongle
	% meiyoujisuangangdu zhen ershi cancha 
	%fem buxuyao zongshinneng (offline varitional operation)
	% online and offline
	% elegant deep learning tensor style
	% 
	%In this section, since input-output shi deep learning changjiande, we first introduce
	%
	%energy implementation-- variational operation implementation
	%The framework of 
	%In the last section, functional is approximated by Eq. \ref{eq:discretetotal2}, which does not include stiffness matrix.
	% This is a kind of interpolation,
	In the framework of VOL, we mainly consider two optimization strategies, direct minimization (VOL+DM) and iterative update to optimize Eq. \ref{eq:goal}.
	\begin{itemize}
	\item \emph{Direct minimization.}
	VOL+DM sets the residual norm as the loss function directly, which is similar to many existing domain knowledge embedding methods \cite{raissi_2019_physicsinformed,Gao2021,Ranade2021}. But according to our experimental results, see section \ref{sec:ddos}, VOL+DM performs not as great as the iterative update. 
	\item \emph{Iterative update.}
	In this Article, we consider two common cases of the iterative update, i.e., the steepest decent update (SD) and the conjugate gradient update (CG). They are wrapped as $Opt$ used in line 7 of Algorithm \ref{alg:vol}. Instead of running the iterative method until it converges, we only use a fixed number of update steps of the iterative method in one forward propagation, which is more computationally efficient and proves to be effective in our experiments (See section \ref{sec:sizescaling}, where VOL+CG(2) shows a power law). SD and CG with $n$ steps update are shown in Algorithm \ref{alg:steepestdecent} and Algorithm \ref{alg:CG} respectively. They take all of the residual tensors of $j$th batch $\mathbf{R}_j^{bs}$, the prediction of current batch $\mathbf{a}_j^{bs}$ and input parameter $\mathbf{U}_j^{bs}$ as input, and return the update of current prediction $\Delta\mathbf{a}_j^{bs}$. To calculate $\mathbf{R}^\mathrm{T} \mathbf{K} \mathbf{R} $ in Algorithm \ref{alg:steepestdecent} and $\mathbf{p}^\mathrm{T} \mathbf{K} \mathbf{p} $ in Algorithm \ref{alg:CG}, it is also unnecessary to obtain $\mathbf{K}$. All we need to do is to use Ritz approach to $\mathbf{R}$, even with a more simplified calculation than Eq. \ref{eq:discretetotal2}:
	\begin{equation}\label{eq:RtKR}
		\begin{gathered}
			\mathbf{R}^\mathrm{T} \mathbf{K} \mathbf{R} = \sum_e \sum_{l=1}^{n_g} H_l \mathbf{R}^{e \mathrm{T}} \mathbf{B}_l^{\mathrm{T}} \mathbf{D}_l \mathbf{B}_l \mathbf{R}^{e}  \left | \mathbf{J}_l^e\right |  \\
			\mathbf{p}^\mathrm{T} \mathbf{K} \mathbf{p} = \sum_e \sum_{l=1}^{n_g} H_l \mathbf{p}^{e \mathrm{T}} \mathbf{B}_l^{\mathrm{T}} \mathbf{D}_l \mathbf{B}_l \mathbf{p}^{e}  \left | \mathbf{J}_l^e\right |
		\end{gathered}
	\end{equation}
	
	Similarly, the forward-backward propagation loop between $\mathbf{a}$ and $\widetilde{\Pi}_{temp}$ can also be constructed by Ritz approach to derive the residual tensor of the next update step $\mathbf{R}$ in line 10 of Algorithm \ref{alg:steepestdecent}, and between $\mathbf{p}$ and $\mathbf{p}^\mathrm{T} \mathbf{K} \mathbf{p}$ to derive $\mathbf{Kp}$ in line 9 of Algorithm \ref{alg:CG}. Besides, we can also choose Galerkin approach to operate on $\mathbf{a}$, $\mathbf{p}$ to derive $\mathbf{R}$ and $\mathbf{Kp}$. For efficiency, we always use Galerkin approach to calculate matrix-vector products. Then the same mask operation is also imposed on $\mathbf{R}$ and $\mathbf{K} \mathbf{p}$ respectively. 
	
	% Considering more steps of iterative methods will increase the computational cost, for all experiments in this work, when we use SD, we use only one step of SD in one inner loop in VOL, and when we use CG, we use two steps of CG in one inner loop in VOL.
	
	With the update of current prediction $\Delta\mathbf{a}_j^{bs}$, we can manage to update the parameter $\bm{\vartheta  }$ of the neural operator module. Here, we adopt a simple way to derive the update of $\bm{\vartheta  }$. First, we assume $\mathbf{a}_j^{bs}+\Delta\mathbf{a}_j^{bs}$ to be a "provisional label" of the current batch. Then, we can define the loss metric that corresponds to the provisional label $\hat{\mathbf{a}_j^{bs}}=\mathbf{a}_j^{bs}+\Delta\mathbf{a}_j^{bs}$ and the current prediction $\mathbf{a}_j^{bs}$, and we choose to use sum of squares error (SSE) as the loss metric:
	\begin{equation}\label{eq:lossmetric}
		\begin{gathered}
			\mathcal{L}_{\text{SSE}}=\frac{1}{2} \verb|sum|\left( \verb|square|\left( \hat{\mathbf{a}_j^{bs}} -\mathbf{a}_j^{bs}\right)\right) 
		\end{gathered}
	\end{equation}
	
	Then in line 9 of Algorithm \ref{alg:vol}, The gradient of $\bm{\vartheta}$ is calculated with the chain rule\footnote{For convenience, we consider the all trainable parameters $\bm{\vartheta}$ to form a vector.}, which is easy to implement with \verb|Function| class in \verb|torch.autograd| module in Pytorch \cite{PyTorch2019}, then, the update of $\bm{\vartheta}$ is calculated with $Opt_\text{net}$
	\begin{equation}\label{eq:updatetheta}
		\begin{gathered}
			\bm{\vartheta  }=\bm{\vartheta  } - \hat{\eta}  \cdot Opt_\text{net}\left(\verb|einsum| \left( \verb|'| \cdots \verb|t,| \cdots \verb|->t',| \verb|grad|\left( \mathbf{a}_{j}^{bs},\bm{\vartheta}\right) \verb|,|  \verb|grad|\left(\mathcal{L}_{\text{SSE}}, \mathbf{a}_{j}^{bs}\right)  \right) \right)
			\\= \bm{\vartheta  } + \hat{\eta}  \cdot Opt_\text{net} \left(\verb|einsum| \left( \verb|'| \cdots \verb|t,| \cdots \verb|->t',| \verb|grad|\left( \mathbf{a}_{j}^{bs},\bm{\vartheta}\right) \verb|,| \Delta \mathbf{a}_{j}^{bs} \right) \right) 
		\end{gathered}
	\end{equation}
	
\end{itemize}

	% $w_1, w_2, w_3, w_4$ are pre-computed and fixed weight of convolution filters, according to the values of interpolation functions and their derivatives.
	% The length, width and height of the 3D convolution filters and the width and height of the 2D convolution filters should be determined according to the shape of the element of computational mesh, and the number of the filters equals to the number of Gauss points in one element.
	% (e.g., element material property tensor, element thermal conductivity tensor) with Eq. \ref{eq:discretetotal2}.
	% Consider Jacobian effect on the integrand. A Jacobian tensor is constructed, with the same shape as the integrand, and then the element-wise product of the integrand tensor and the Jacobian tensor is calculated. 
	% and Gaussian quadrature to get tensor $\left \{ \Pi^e \right \} $ that has shape 1$\times$3$\times$3, and sum up all elements of $\left \{ \Pi^e \right \} $ to obtain $\Pi$.

	\begin{algorithm}[H]
		\begin{algorithmic}[1]
			\REQUIRE residual tensor $\mathbf{R}$, current node solution $\mathbf{a}$, input parameter $\mathbf{U}$, mask operation $Mask$
			\STATE $\mathbf{R}=-\mathbf{R}$
			\STATE $ \Delta \mathbf{a}=\mathbf{0}$
			\FOR{$1\leq i \leq n $} 
			\STATE Calculate $\mathbf{R}^\mathrm{T} \mathbf{K} \mathbf{R} $ corresponding to $ \left( \mathbf{U},\mathbf{R} \right)  $
			\STATE $\alpha=\frac{\mathbf{R}^\mathrm{T} \mathbf{R}}{\mathbf{R}^\mathrm{T} \mathbf{K} \mathbf{R}}$
			\STATE $ \Delta \mathbf{a}$+=$\alpha \mathbf{R} $
			\IF {$n>1$}
			\STATE $\mathbf{a}$+=$\alpha \mathbf{R}$
			\STATE Get $\mathbf{R}$ with Ritz approach or Galerkin approach
			\STATE $\mathbf{R}= Mask\left( \mathbf{R} \right) $		
			\STATE $\mathbf{R}=-\mathbf{R}$
			\ENDIF
			\ENDFOR
			\ENSURE $\Delta \mathbf{a}$
		\end{algorithmic}
		\caption{\textbf{Steepest descent} ($n$ \textbf{steps})}
		\label{alg:steepestdecent}
	\end{algorithm}

	\begin{algorithm}[H]
		\begin{algorithmic}[1]
			\REQUIRE residual tensor $\mathbf{R}$, current node solution $\mathbf{a}$, input parameter $\mathbf{U}$, mask operation $Mask$
			\STATE $\mathbf{R}=-\mathbf{R}$
			\STATE $\mathbf{p}=\mathbf{R}$
			\STATE $ \Delta \mathbf{a}=\mathbf{0}$
			\FOR{$1\leq i \leq n $} 
			\STATE Calculate $\mathbf{p}^\mathrm{T} \mathbf{K} \mathbf{p} $ corresponding to $ \left( \mathbf{U},\mathbf{p} \right)  $
			\STATE $\alpha=\frac{\mathbf{R}^\mathrm{T} \mathbf{R}}{\mathbf{p}^\mathrm{T} \mathbf{K} \mathbf{p}}$
			\STATE $ \Delta \mathbf{a}$+=$\alpha \mathbf{p} $
			
			\IF {$n>1$}
			\STATE Get $\mathbf{Kp}$ with Ritz approach or Galerkin approach  
			\STATE $\mathbf{Kp}= Mask\left( \mathbf{Kp} \right) $  		
			\STATE $ \mathbf{R}_{\text{new}}=\mathbf{R}-\alpha \mathbf{Kp}$
			\STATE $\beta= \frac{\mathbf{R}_{\text{new}}^\mathrm{T} \mathbf{R}_{\text{new}}}{\mathbf{R}^\mathrm{T} \mathbf{R}}$
			\STATE $\mathbf{R}=\mathbf{R}_{\text{new}}$
			\STATE $ \mathbf{p}=\mathbf{R}+\beta \mathbf{p}$
			\ENDIF
			\ENDFOR
			\ENSURE $\Delta \mathbf{a}$
		\end{algorithmic}
		\caption{\textbf{Conjugate gradient decent} ($n$ \textbf{steps})}
		\label{alg:CG}
	\end{algorithm}
	
% 	\begin{algorithm}[H]
% 	\begin{algorithmic}[1]
% 		\REQUIRE residual tensor $\mathbf{R}$, current node solution $\mathbf{a}$, input parameter $\mathbf{U}$, mask operation $Mask$
% 		\STATE $\mathbf{R}=-\mathbf{R}$
% 		\STATE $\mathbf{p}=\mathbf{R}$
% 		\STATE $ \Delta \mathbf{a}=\mathbf{0}$
% 		\FOR{$1\leq i \leq n $} 
% 		\STATE Calculate $\mathbf{p}^\mathrm{T} \mathbf{K} \mathbf{p} $ corresponding to $ \left( \mathbf{U},\mathbf{p} \right)  $
% 		\STATE $\alpha=\frac{\mathbf{R}^\mathrm{T} \mathbf{R}}{\mathbf{p}^\mathrm{T} \mathbf{K} \mathbf{p}}$
% 		\STATE $ \Delta \mathbf{a}$+=$\alpha \mathbf{p} $
		
% 		\IF {$n>1$}
% 		\STATE Get $\mathbf{Kp}$ with Ritz approach or Galerkin approach  
% 		\STATE $\mathbf{Kp}= Mask\left( \mathbf{Kp} \right) $  		
% 		\STATE $ \mathbf{R}_{\text{new}}=\mathbf{R}-\alpha \mathbf{Kp}$
% 		\STATE $\beta= \frac{\mathbf{R}_{\text{new}}^\mathrm{T} \mathbf{R}_{\text{new}}}{\mathbf{R}^\mathrm{T} \mathbf{R}}$
% 		\STATE $\mathbf{R}=\mathbf{R}_{\text{new}}$
% 		\STATE $ \mathbf{p}=\mathbf{R}+\beta \mathbf{p}$
% 		\ENDIF
% 		\ENDFOR
% 		\ENSURE $\Delta \mathbf{a}$
% 	\end{algorithmic}
% 	\caption{\textbf{Generalized minimal residual method} ($n$ \textbf{steps})}
% 	\label{alg:GMRES}
% \end{algorithm}	

	\section{Conclusion and outlook}
	\label{sec:conc}

	Both the conventional solver and the data-driven surrogate modeling have their own merits and shortcomings. The conventional solvers, which are based on the domain knowledge, can give reliable solutions to a wide variety of PDEs, and the conventional solvers are affordable when dealing with a single instance of PDEs or small number of samples, but they often have low efficiency and bring a heavy computational burden when the mass sampling of parameters of PDEs is required, because they usually only solve a single parameter each time. On the contrary, the data-driven surrogate modeling can give reasonable prediction of solutions to a range of parameters at a fast inference speed. However, as mentioned in Introduction (see section \ref{sec:introduction}), the data-driven surrogate modeling needs a data preparation stage and a model training stage, which are isolated from each other. Besides, the access to substantial labeled data brings also quite a computational burden to the data-driven surrogate modeling. In machine learning field, more efficient training and less use of labels mean lower carbon, lower dataset costs and shorter experimental cycles. In this work, a novel data-efficient paradigm that has the merits of both and complements the shortcomings has been proposed, which we refer to as the variational operator learning (VOL). Our proposed VOL is part of an important effort to reduce the carbon footprint of research in the field of machine learning for PDEs and the numerical simulation. The proposed VOL achieves matrix-free approximation of system functional and residual with Ritz and Galerkin approaches. Direct residual minimization and iterative update, as two optimization strategies, are then proposed in VOL to learn PDEs' solution operators with a label-free training set. We have conducted various experiments on the variable heat source problem, the Darcy flow problem, and variable stiffness elasticity problem with two cases to demonstrate the effectiveness of VOL. Scaling experiments show test errors of VOL also follows a power law like LLMs \cite{kaplan_2020_scaling,maloney_2022_a}, and is able to learn solution operators with satisfactory results provided with enough cheap unlabeled data. Resolution experiments show VOL can learn solution operators largely efficiently when the solution field is discretized at different resolutions. We then design comparative experiments to verify generalization benefits of VOL, where we observe VOL has more superiority than classical iteration methods as the resolution increases. However, we also observe that, compared to classical iterative methods, the convergence speed is not significantly improved by the current VOL algorithm, and the test errors of VOL have little improvement with larger number of update steps. Though we focus on training a neural operator module with iterative updates for the real-time inference, rather than accelerating the iterative solvers in the current work, it is still necessary to utilize techniques like preconditioners \cite{kechen_2005_matrix}, multigrid methods \cite{ulrichtrottenberg_2001_multigrid}, neural-accelerated solvers \cite{stanziola2021helmholtz,CHEN2022110996}, etc., to accelerate iterations. We then conduct comparative experiments for optimization strategies of VOL and the conventional data-driven strategy, results show the smaller residual norm does not refer to a smaller test error. While VOL achieves a smaller residual norm, it has a slightly larger test error than the data-driven strategy with the same training settings. 

	There are several aspects to be explored for the proposed VOL in the future work. Although VOL shows promise for label-free training of neural operators, it uses standard convolution operations, which limits its application on solution domains of arbitrary shapes. Graph operations and meshless methods will be considered in our future work to enhance the flexibility of the proposed VOL to handle solution domains with arbitrary shapes. In this Article, we mainly consider heat transfer, Darcy flow and elasticity physics at their steady-state cases, and in the future work, we will consider VOL on more complex physics phenomena, such as fluid dynamics, electromagnetics, and quantum mechanics. More iteration techniques and neural architectures will also be considered in VOL, such as generalized minimum residual method \cite{ysaad_2003_iterative}, multigrid algorithms \cite{ulrichtrottenberg_2001_multigrid,xu_2017_algebraic}, factorized Fourier neural operators \cite{tran2023factorized} and geometry-informed neural operators \cite{li_2023_geometryinformed}.

	\section{Acknowledgments}

	This work was supported by National Key Research and Development Program of China (2021YFF0306404), and National Natural Science Foundation of China (U21A20429 and U11772078).

	\section{Author contributions}
	T.X. and P.H. contributed to the original idea and design of the research. T.X. and D.L generated the datasets. T.X were responsible for software, data curation and formal analysis. P.H. acquired funding, and was responsible for the administration, resources and supervision of the project. All authors wrote, reviewed and edited the manuscript.

	\bibliographystyle{plain}
	\bibliography{main}

	\appendix
	\setcounter{figure}{0}
	\renewcommand{\figurename}{Extended Data Figure}

	\begin{figure}[htbp]
		\centering
		\includegraphics[width=0.4\textwidth]{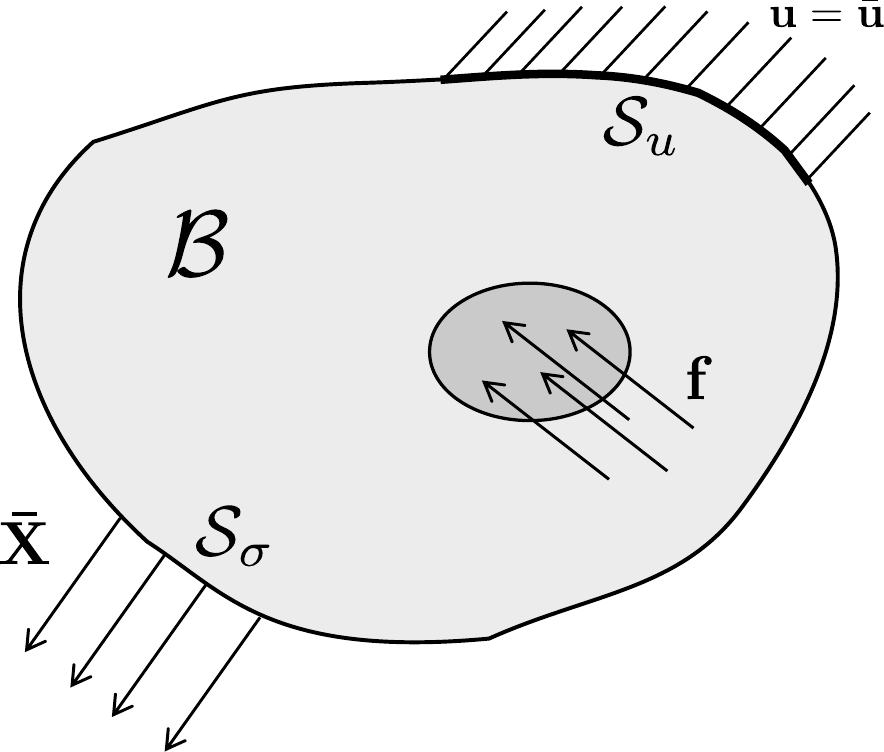}
		\caption{A linear elastic body with body forces and boundary conditions.}
		\label{fig:elasticbody}
	\end{figure}

	\begin{figure}[htbp]
		\centering
		\includegraphics[width=1\textwidth]{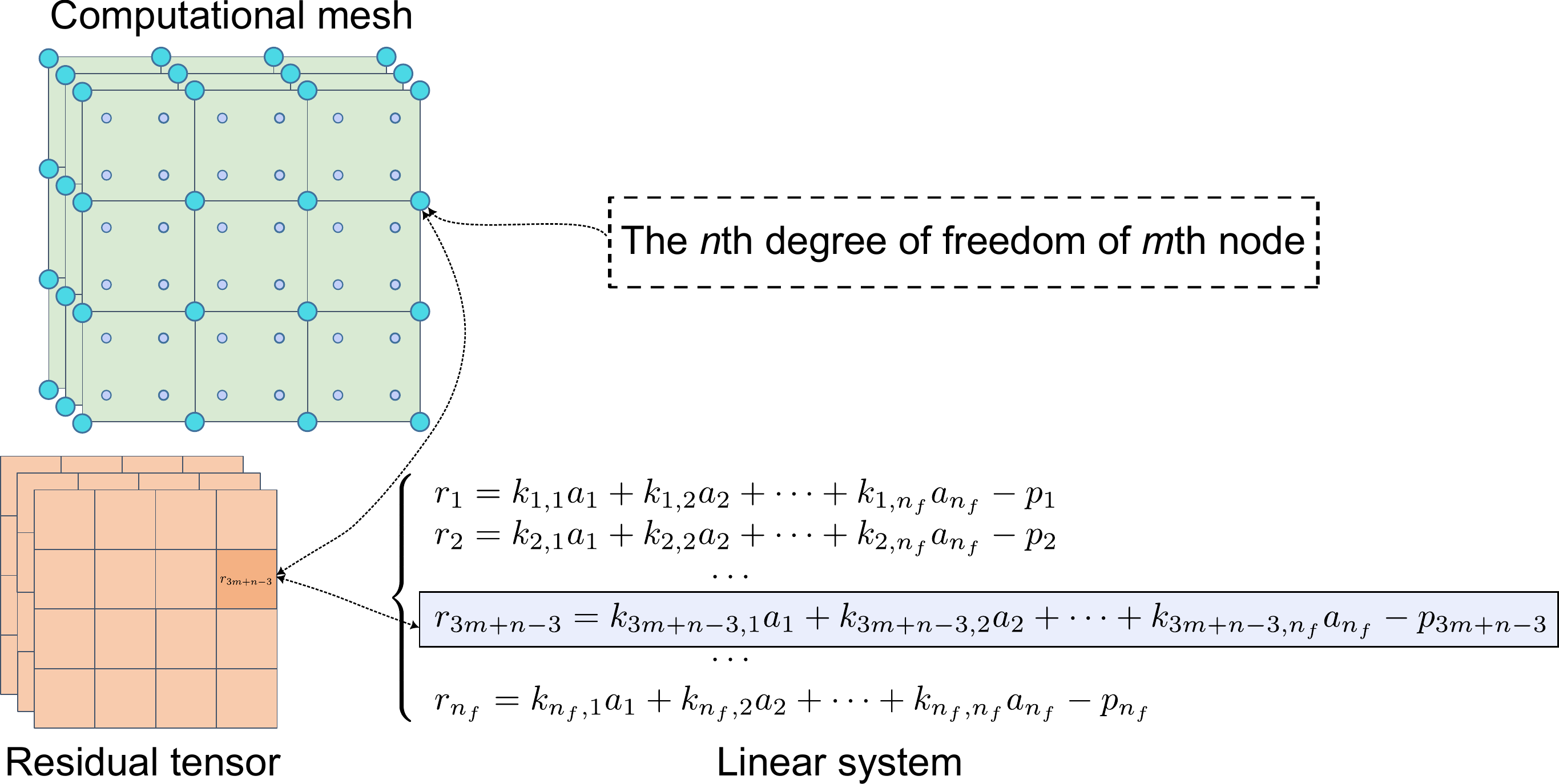}
		\caption{\textbf{The relationship between the computational mesh, the residual tensor and the linear system.} Suppose that every node of the computational mesh has three degrees of freedom. The element $r_{3m+n-3}$ of the residual tensor is just the residual of the $\left( 3m+n-3\right) $th equation of the corresponding linear system. }
		\label{fig:residual--equation}
	\end{figure}

	\begin{figure}[htbp]
		\centering
		\includegraphics[width=1\textwidth]{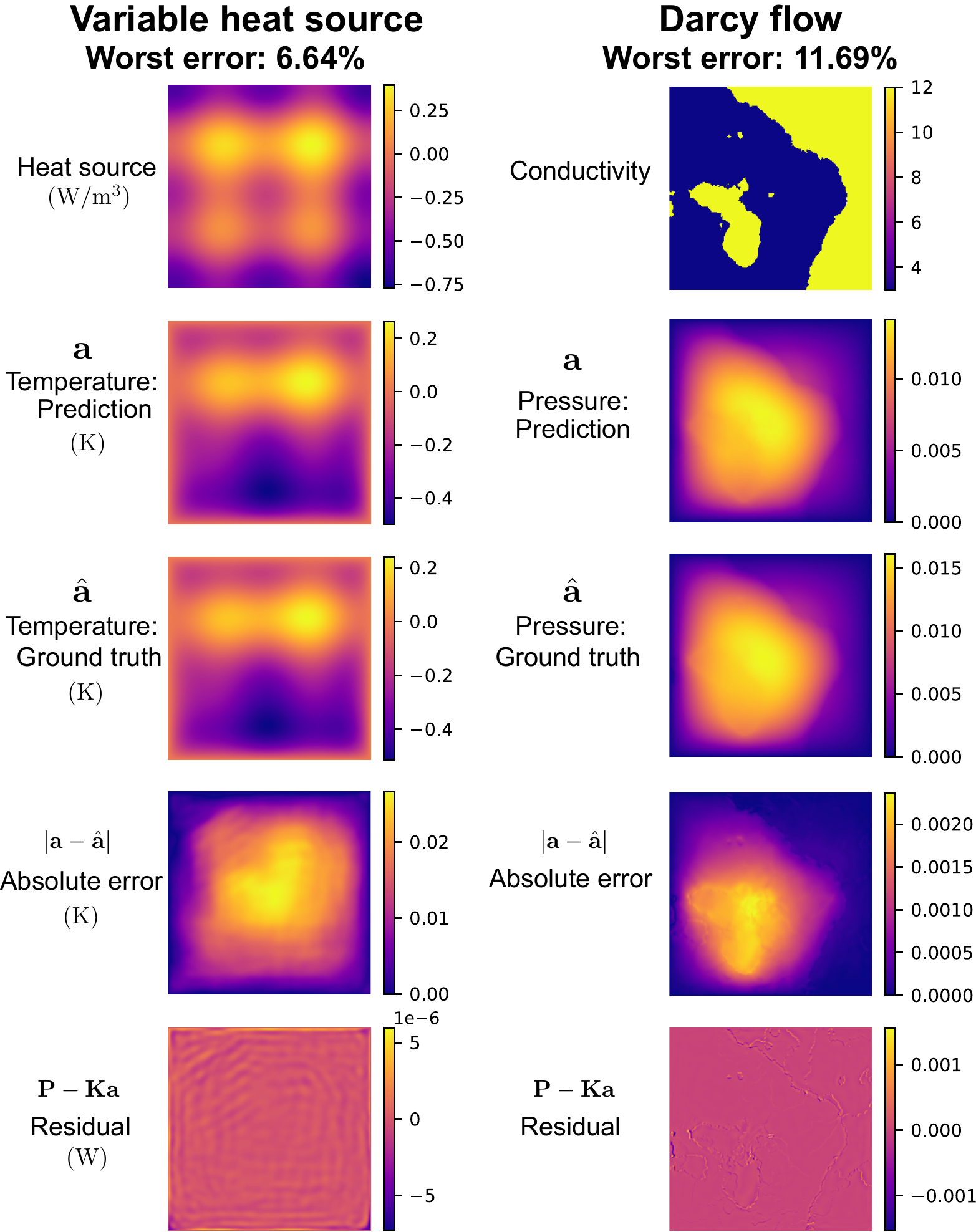}
		\caption{\textbf{Worst samples of variable heat source problem and darcy problem in scaling experiments.} ($N_{\text{training set}}=10000$, $N_{\text{test set}}=2000$).}
		\label{fig:worst12}
	\end{figure}
	
	% We plot the worst sample of variable heat source problem with its heat source ($\text{W}/\text{m}^3$), temperature prediction, temperature ground truth, temperature absolute error (K), and residual (W) and plot the worst sample of the darcy problem with its conductivity, pressure prediction, pressure prediction, pressure absolute error, residual in scaling experiments.

	\begin{figure}[htbp]
		\centering
		\includegraphics[width=1\textwidth]{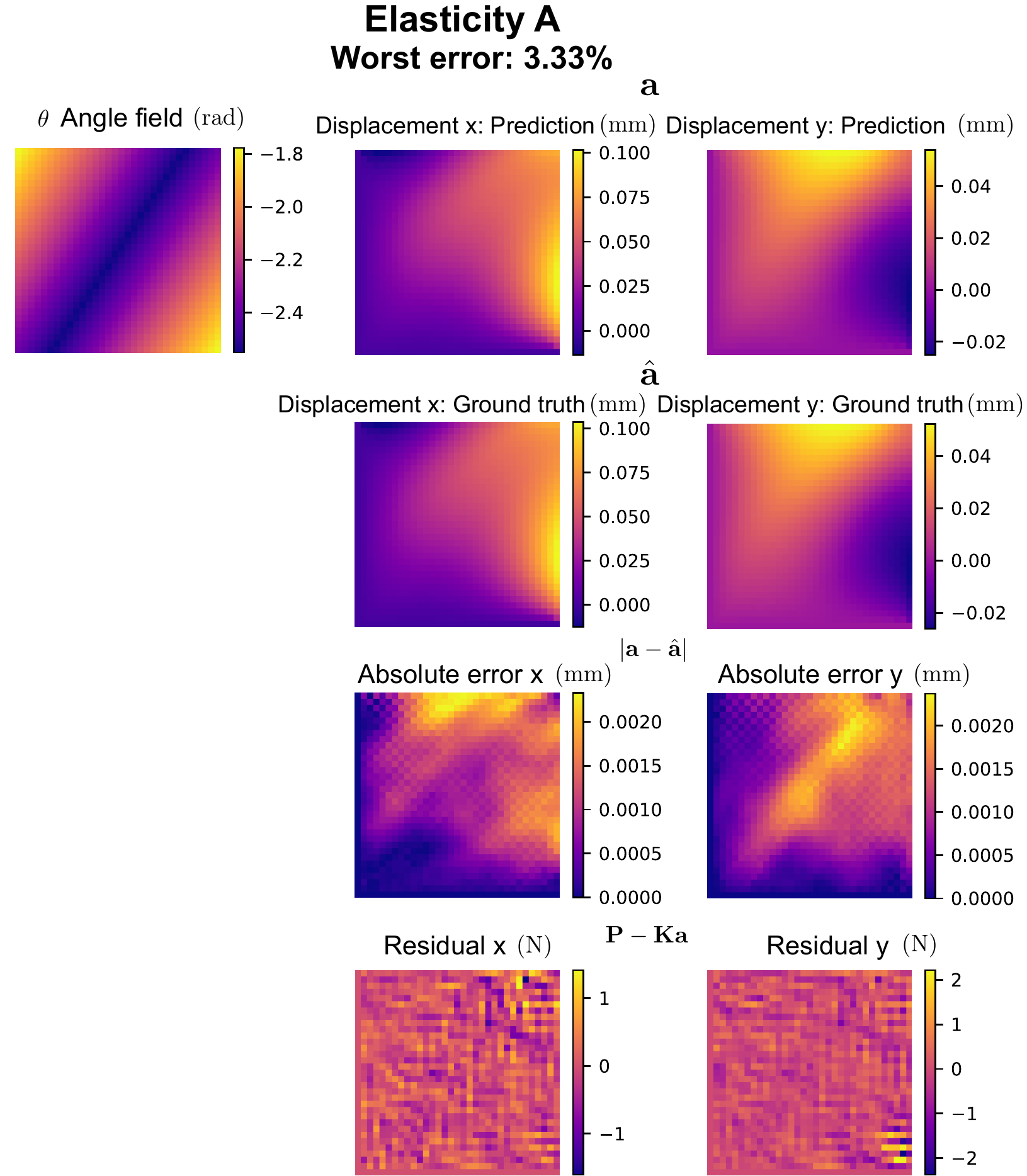}
		\caption{\textbf{Worst samples of elasticity A problem in scaling experiments.} ($N_{\text{training set}}=10000$, $N_{\text{test set}}=2000$).}
		\label{fig:worst3}
	\end{figure}

	\begin{figure}[htbp]
		\centering
		\includegraphics[width=1\textwidth]{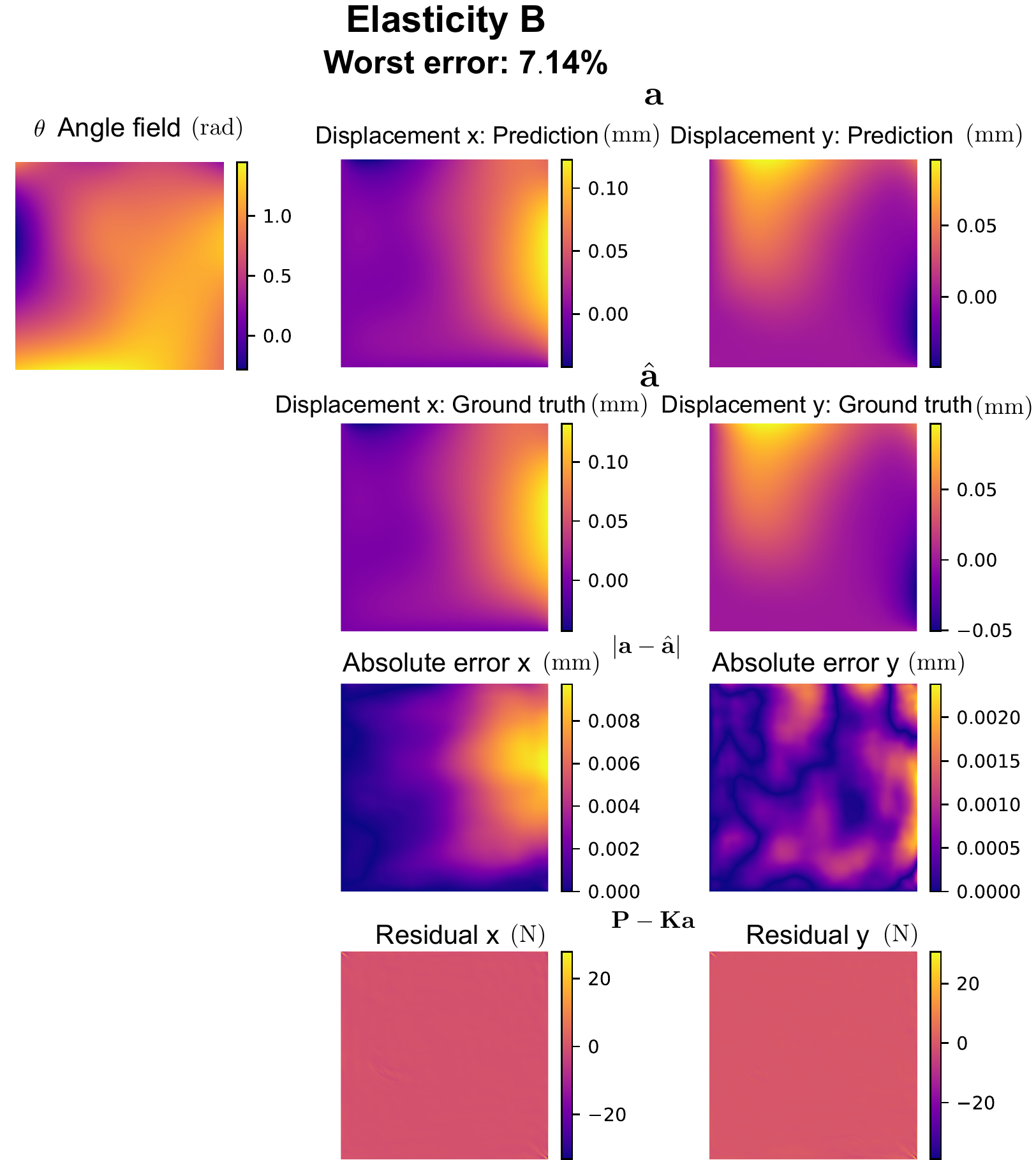}
		\caption{\textbf{Worst samples of elasticity B problem in scaling experiments.} ($N_{\text{training set}}=10000$, $N_{\text{test set}}=2000$).}
		\label{fig:worst4}
	\end{figure}

\end{document}